\documentclass[10pt,journal,compsoc]{IEEEtran}
\ifCLASSOPTIONcompsoc
  \usepackage[nocompress]{cite}
\else
  \usepackage{cite}
\fi
\ifCLASSINFOpdf
\else
\fi
\usepackage{amsfonts}
\usepackage{graphicx}
\usepackage{epstopdf}
\usepackage{algorithm}
\usepackage{amsmath}
\usepackage{amssymb}
\usepackage{multirow,array}
\usepackage{longtable}
\usepackage{enumerate}
\usepackage{mathrsfs}
\usepackage{threeparttable}
\usepackage{ragged2e}

\usepackage{algorithmicx}
\usepackage{lipsum}
\usepackage{xifthen}
\usepackage{needspace}
\usepackage{hyperref}

\newcounter{myalg}
\newenvironment{myalg}[1][]
  {
    \needspace{2\baselineskip}
    \noindent \rule{\linewidth}{1pt} \endgraf
    \refstepcounter{myalg}
    \centering \textsc{Algorithm}~\themyalg
    \ifthenelse{\isempty{#1}}{}{:\ #1}
  }{
  \noindent \rule{\linewidth}{1pt}
  }

\begin{document}
\title{A Unified Joint Matrix Factorization Framework for Data Integration}
\author{Lihua~Zhang
        and~Shihua~Zhang
\IEEEcompsocitemizethanks{\IEEEcompsocthanksitem Lihua Zhang and Shihua Zhang are with the National Center for Mathematics and Interdisciplinary Sciences, Academy of Mathematics and Systems Science, CAS, Beijing, China, 100190, and School of Mathematics Sciences, University of Chinese Academy of Sciences, Beijing 100049, China. \protect\\ Email: zsh@amss.ac.cn.}}

\IEEEtitleabstractindextext{
\begin{abstract}
\justifying
Nonnegative matrix factorization (NMF) is a powerful tool in data exploratory analysis by discovering the hidden features and part-based patterns from high-dimensional data. NMF and its variants have been successfully applied into diverse fields such as pattern recognition, signal processing, data mining, bioinformatics and so on. Recently, NMF has been extended to analyze multiple matrices simultaneously. However, a unified framework is still lacking. In this paper, we introduce a sparse multiple relationship data regularized joint matrix factorization (JMF) framework and two adapted prediction models for pattern recognition and data integration. Next, we present four update algorithms to solve this framework. The merits and demerits of these algorithms are systematically explored. Furthermore, extensive computational experiments using both synthetic data and real data demonstrate the effectiveness of JMF framework and related algorithms on pattern recognition and data mining.
\end{abstract}

\begin{IEEEkeywords}
non-negative matrix factorization, joint matrix factorization, data integration, network-regularized constraint, pattern recognition, bioinformatics.
\end{IEEEkeywords}}

\maketitle

\IEEEdisplaynontitleabstractindextext
\IEEEpeerreviewmaketitle

\IEEEraisesectionheading{\section{Introduction}\label{sec:introduction}}
\IEEEPARstart{N}{onnegative} matrix factorization (NMF) is a powerful matrix factorization technique which typically decomposes a nonnegative data matrix into the product of two low-rank nonnegative matrices \cite{paatero1994positive,lee1999learning}. NMF was first introduced by Paatero and Tapper (1994) and has become an active area with much progress both in theory and in practice since the work by Lee and Seung (1999). NMF and its variants have been recognized as valuable exploratory
analysis tools. They have been successfully applied into many fields including signal processing, data mining, pattern recognition, bioinformatics and so on \cite{berry2007algorithms,wang2013nonnegative}.

NMF has been shown to be able to generate sparse and part-based representation of data \cite{lee1999learning}. In other words, the factorization allows us to easily identify meaningful sub-structures underlying the data. In the past decade, a number of variants have been proposed by incorporating various kinds of regularized terms including discriminative constraints \cite{zafeiriou2006exploiting}, network-regularized or locality-preserving constraints \cite{Cai2008non,zhou2009local}, sparsity constraints \cite{hoyer2004non,Kim2007sparse}, orthogonality constraints \cite{Ding2006orthogonal} and others \cite{zhi2011graph}.

However, the typical NMF and its variants in its present form can only be applied to one matrix containing just one type of variables.
Large amounts of multi-view data describing the same set of objects can be available now. Thus, data integration methods are urgently needed. Recently, joint matrix factorization based data integration methods have been proposed for pattern recognition and data mining among pairwise or multi-view data matrices. For example, Greene and Cunninghan proposed an integration model based on matrix factorization (IMF) to learn the embedded underlying clustering structures across multiple views. IMF is a late integration strategy, which fuses the clustering solutions of each individual view for further analysis \cite{greene2009matrix}. Zhang \emph{et al.} (2012) proposed a joint nonnegative matrix factorization (jNMF) to decompose a number of data matrices $X_I$ which share the same row dimension into a common basis matrix $W$ and different coefficient matrices $H_I$, such that $X_I \approx WH_I$ by minimizing $\sum_{I}\|X_I - WH_I\|_F^2$ \cite{zhang2012discovery}. This simultaneous factorization can not only detect the underlying part-based patterns in each matrix, but also reveal the potential connections between patterns of different matrices. A further network-regularized version has also been proposed and applied in bioinformatics \cite{zhang2011novel}. Liu \emph{et al.} (2013) proposed a multi-view clustering method, which factorizes individual matrices simultaneous and requires the coefficient matrices learnt from various views to be approximately common \cite{liu2013multi}. Specifically, it is defined as follows,
\begin{equation*}\label{eqSNMFCA}
  \mbox{\small$\displaystyle
  \begin{aligned}
     \mbox{min}\quad & \sum_{I}\|X_I - W_IH_I\|_F^2 + \sum_{I}\lambda_{I}\|W_I - W^*\|_F^2 \\
      \mbox{s.t.}\quad    & \|(H_I)_{i,\cdot}\|_1=1, \forall i \,\mbox{and}\, W_I\geq 0, H_I\geq 0, W^*\geq 0,
   \end{aligned}
$}
\end{equation*}
where $\lambda_{I}$ is a parameter to tune the relative weight among different views as well the two terms.
Zitnik and Zupan (2015) proposed a data fusion approach with penalized matrix tri-factorization (DFMF) for simultaneously factorizing multiple relationship matrices $R_{ij}$ in one framework \cite{vzitnik2015data}. They also considered to incorporate the must-link and cannot-link constraints within each data type into the DFMF model as follows,
\begin{equation*}\label{eqSNMFCA}
  \mbox{\small$\displaystyle
  \begin{aligned}
     \mbox{min}\quad & \sum\limits_{{{R}_{ij}}\in R}{{{\left\| {{R}_{ij}}-{{G}_{i}}{{S}_{ij}}G_{j}^{T} \right\|}^{2}}}+\sum\limits_{t=1}^{\max ({{t}_{i}})}{Tr({{G}^{T}}{{\Theta }^{(t)}}G)} \\
      \mbox{s.t.}\quad    &  G\geq 0,
   \end{aligned}
$}
\end{equation*}
where $G = \mbox{diag}(G_{1},G_{2},\dots ,G_{r})$, ${\Theta}^{(t)}=\mbox{diag}({\Theta_{1}}^{(t)},{\Theta_{2}}^{(t)},\dots ,{\Theta_{r}}^{(t)})$, $t \in \{1,2,\dots ,{\mbox{max}_{i}t_{i}}\}$, $t_{i}$ is the number of data sources for the $i$th object type. $R_{ij}$ represents the relationship data matrix between the $i$th and the $j$th object type (between constraint). DFMF decomposes it into $G_{i}$, $G_{j}$ and $S_{ij}$ constrained by ${\Theta_{i}}^{(t)}$ (within constraint), which provides relations between objects of the $i$th object type.
This method well exploits the abstracted relationship data, but ignores the sample-specific information of data. In image science, Jing \emph{et al.} (2012) adopted a supervised joint matrix factorization model to learn latent basis by factorizing both the region-image matrix $X_1\in R^{n\times m}$ and the annotation-image matrix $B \in R^{n\times p}$ simultaneously and incorporating the label information $Y$ (where $Y_i$ indicates the label index of the $i$th image) \cite{jing2012snmfca}. This supervised model for image classification and annotation (SNMFCA) is formulated as follows,
\begin{equation*}\label{eqSNMFCA}
  \mbox{\small$\displaystyle
  \begin{aligned}
     \mbox{min}\quad & \frac{\lambda}{2}\|X_{1}-W_{1}H\|_F^{2} + \frac{1-\lambda}{2}\|X_{2}-W_{2}H\|_F^{2}+ \frac{\eta}{2}Tr(H \Theta H^T) \\
      \mbox{s.t.}\quad    & W_1\geq 0, W_2\geq 0, H\geq 0,
   \end{aligned}
$}
\end{equation*}
where $ \Theta \in R^{n\times n}$ with $\Theta_{i,j} = 1$ if $Y_i \neq Y_j$ and 0 otherwise. Obviously, the SNMFCA aims to determine the latent basis with known class information. However, this model does not consider the the must-link and cannot-link constraints within each data type and those between data types.

Recently, based on the jNMF \cite{zhang2012discovery}, Stra\v{z}ar \emph{et al.} (2016) proposed an integrative orthogonality-regularized nonnegative matrix factorization (iONMF) to predict protein-RNA interactions. iONMF was an extension of jNMF by integrating multiple types of data with orthogonality regularization on the basis matrix \cite{stravzar2016orthogonal}. This model learns the coefficients matrices from the training dataset, and the basis matrix from the testing dataset, and then predicts the interaction matrix. However, both jNMF and iONMF were originally solved by a multiplicative update method, which might be limited by its slow convergence or even non-convergence issues.

In this paper, we first generalize and introduce a unified joint matrix factorization framework (JMF) based on the classical NMF and jNMF for pattern recognition and data mining by integrating multi-view data $X_I$ on the same objects and must-link and cannot-link constraints within and between any two data. In addition, sparsity constraints are also considered. We adopt four update algorithms including multiplicative update algorithm (MUR), projected gradient method (PG), Nesterov's optimal gradient method (Ne), and a novel proximal alternating nonnegative least squares algorithm (PANLS) for solving JMF. 
Then, the JMF is extended to two types of prediction models with one based on the basis matrix $W$ and another based on the coefficients matrices $H_{I}$ ($I = 1, 2,\dots , N$). Finally, we demonstrate the effectiveness of this framework both in revealing object-specific multiple-view hidden patterns and prediction performance through extensive computational experiments.

Compared with existing NMF techniques for pattern recognition and data integration, JMF has the following characteristics:
\begin{enumerate}[(i)]
  \item JMF can model multi-view data as well as must-links/cannot-links simultaneously for recognizing object-specific and multi-view associated patterns.
  \item Must-links and cannot links within and between some views can be completely missing, and each within-view or between-view type can be associated with multiple constraint matrices.
  \item JMF can be solved with diverse update algorithms, among which PANLS is a representative one for solving JMF with competitive performance in terms of computational accuracy and efficiency.
\end{enumerate}

The rest of the paper is organized as follows. In section \ref{Problem}, we describe the formulation of JMF. In section \ref{Algorithm}, we present four update methods to solve JMF. In section \ref{Prediction}, we propose two prediction models based on JMF. In section \ref{Experiments}, we illustrated the experimental results on both synthetic and real datasets. At last, we summarize this study in section \ref{Conclusion}.

\section{Problem formulation}\label{Problem}
Given two nonnegative matrices $X_1$ and $X_2$ with size of $m\times n_1$ and $m\times n_2$, the networked relationship represented by two adjacency matrices $\Theta_1$ and $\Theta_2$ with size of $n_1\times n_1$ and $n_2\times n_2$ and the between networked relationship represented by a bipartite adjacency matrix $R_{12}$ of size $n_1\times n_2$. In our application, our assumption is that the two matrices $X_1$ and $X_2$ are two different kinds of descriptions of the same set of objects, the networked relationship $\Theta_1$, $\Theta_2$ and $R_{12}$ are described as prior knowledge about the features. The goal of this study is to find a reduced representation by incorporating all the data we have now.

To achieve the ultimate goal in one framework, we incorporate three components into the objective function. The first one considers the parts-based data representation of two matrices $X_1$ and $X_2$. The second and third ones consider the networked relationship $\Theta_1$ and $\Theta_2$ of each type of features, and the between networked relationship $R_{12}$ by imposing network regularized constraints, respectively. Finally, we consider to incorporate sparsity constraints to get a sparse solution.

\subsection{NMF and its variants}
Non-negative matrix factorization (NMF) problem is a matrix factorization model which uses two low-rank non-negative matrices, i.e., one basis matrix and one coefficient matrix, to reconstruct the original data matrix \cite{paatero1994positive,lee1999learning}. Its objective
function is
$$
\underset{W,H\geq0}{\text{min}}
\|X - WH\|_F^2,
$$
where $W$ and $H$ are the basis matrix and coefficient matrix with size of $m \times r$ and $r \times n$ respectively, and $\|\cdot \|_F$ is the Frobenius norm of a matrix. The non-negativity has been stated that parts are generally combined additively to form a whole; hence, it can be useful for learning part-based representations. Thus, the so-called NMF can be a useful technique to decipher distinct sub-structures for revealing subtle data structure in the underlying data. Several approaches for solving NMF have been discussed in \cite{berry2007algorithms}, and more variants and applications of NMF can refer a recent review paper \cite{wang2013nonnegative}. 

Here our goal is to find the linked patterns among two matrices. We assume that there is one common basis matrix $W$ between matrices $X_1$ and $X_2$. So a joint non-negative matrix representation can be derived by the following optimization problem,
\begin{equation}\label{eqjNMF}
  \begin{aligned}
     \mbox{min}\quad & \sum_{I=1,2}\|X_I - WH_I\|_F^2 \\
      \mbox{s.t.}\quad       & W\geq 0, H_I\geq 0.
   \end{aligned}
\end{equation}
Ideally, the low-dimensional representation (the coefficient matrices) $H_1$ and $H_2$ for the original matrices $X_1$ and $X_2$ derived based on the best approximation can lead to the linked patterns. However, it is unnecessarily accurate due to the incompleteness and noises of the data and other possible factors. In order to improve the accuracy of the patterns, we incorporate the prior networked knowledge on each data object, and bipartite networked knowledge between the data objects $X_1$ and $X_2$.

\subsubsection{Networks Regularized Constraints}
Let $H_1$, $H_2$ denote the low-rank representation of the original data matrices. To decipher the inherent modular structure in a network or say the closeness information of the objects, we assume that adjacent nodes should have similar membership profiles. Therefore, we enforce the must-link constraints by maximizing the following optimization function for $X_1$ (or similarly for $X_2$):
\begin{equation}
\mathcal{O}_1 = \sum_{ij}(\Theta_1)_{ij}(h_i^1)^Th_j^1
              = Tr(H_1\Theta_1H_1^T), \\
\end{equation}
\noindent where $H_1 = [{h_{1}^{1}}, {h_{2}^{1}},\dots ,{h_{n_{1}}^{1}}]$. Similarly, the between relationship information between the two types of objects can also be adopted in the following objective function:
\begin{equation}
\mathcal{O}_2 = \sum_{ij}(R_{12})_{ij}(h_i^1)^Th_j^2
              = Tr(H_1R_{12}H_2^T).
\end{equation}
The motivation behind the proposed network regularized constraints are actually quite straightforward. Note that the solution of the problem defined in Eq. \ref{eqjNMF} is often not unique. We expect to obtain a solution for Eq. \ref{eqjNMF}, which also satisfies the network-regularized constraints well. The
limitations of the previous model and the noisy of the real data lead us to consider an integrative framework for jointly handing feature data and networked data simultaneously.

\subsubsection{Networks-Regularized jNMF}
Here, we incorporate all the data (represented in these five matrices) to discover linked patterns based on $W$, $H_1$ and $H_2$. Specifically, we combine all the above objective functions together to integrate all the five matrices in the following optimization problem:
\begin{equation}
  \mbox{\small$\displaystyle
  \begin{aligned}
     \mbox{min}\quad &  \sum_{I=1,2}\|X_{I}-WH_{I}\|_F^{2}- \lambda_1 \sum_{I=1,2}Tr(H_I \Theta_I H_I^T) \\
       &-\lambda_2 Tr(H_1R_{12}H_2^T) \\
      \mbox{s.t.}\quad    & W\geq 0, H_1\geq 0, H_2\geq 0,
 \end{aligned}
$}
\end{equation}
where the parameters $\lambda_1$ and $\lambda_2$ weigh the link constraints in $\Theta_1$, $\Theta_2$ and $R_{12}$ respectively. And the first term is to
describe the linked patterns between two data matrices by a shared basis or component matrix $W$, the second term $\sum_{I=1,2}Tr(H_I \Theta H_I^T)$ defines the summation of the within-variable constraints that decipher the modular structure in network $\Theta_1$, $\Theta_2$, and the third term $Tr(H_1 R_{12} H_2^T)$ defines the summation of the between-variable constraints which decipher the modular structure in the bipartite network. Here, we can consider the integration of these known networks as graph regularization of the first objective \cite{Cai2008non} or as a semi-supervised learning problem which aims to enforce the must-link constraints into the framework of pattern recognition, where variables with the `must-link' constraint shall be forced into the same pattern. This can facilitate pattern search by significantly narrowing down the large search space and improve the reliability of the identified patterns.

\subsection{A Unified Joint NMF Model (JMF)}
One of the important characteristics of the NMF is that it usually generates sparse representation that allows us to discover parts-based patterns \cite{lee1999learning}. However, several studies have showed that the generation of a parts-based representation by NMF depends on the data and the algorithm \cite{hoyer2004non}. Several approaches have been proposed to explicitly control the degree of sparseness in the $W$ and/or $H$ factors of the NMF
\cite{hoyer2004non,Kim2007sparse}. The idea of imposing $L_1$-norm based constraints for achieving sparse solution has been successfully and comprehensively utilized in various problems \cite{tibshirani1996regression}. We adopt the strategy suggested by \cite{Kim2007sparse}, to make the coefficient matrices $H_1$ and $H_2$ sparse. Thus, the sparse network-regularized jNMF can be formulated as follows:
\begin{equation*} \label{eqS3MNMF}
\mbox{\small$\displaystyle
 \begin{aligned}
     \mbox{min}\quad &  \sum_{I=1,2}\|X_{I}-WH_{I}\|_F^{2}
      - \lambda_1 \sum_{I=1,2} Tr(H_I \Theta_I H_I^T) \\
     & - \lambda_2 Tr(H_1R_{12}H_2^T)  + \gamma_1\|W\|_F^{2}\\
     &+ \gamma_2 \left(\sum_j\|h_{j}^1\|_1^{2}+ \sum_{j'}\|h_{j'}^2\|_1^{2})\right) \\
      \mbox{s.t.}\quad        &  W\geq 0, H_{I}\geq 0,
      \end{aligned}
$}
\end{equation*}
where $h_j^1$ and $h_{j'}^2$ are the $j$th and $j'$th column of $H_1$ and $H_2$ respectively. The first term favors modules with the data profiles, and the second term as well as the third term summarize all the must-link constraints in the first and second profiles, and between the two profiles. The term $\gamma_1\|W\|_F^{2}$ is used to control the scale of matrix $W$, and $\gamma_2 (\sum_j\|h_{j}^1\|_1^{2}+ \sum_{j'}\|h_{j'}^2\|_1^{2})$ encourages the sparsity. The parameter $\gamma_1\geq 0$ suppress the growth of $W$ and $\gamma_2\geq 0$ controls the desired sparsity.

\begin{figure}[htp]
\centerline{\includegraphics[width=0.48\textwidth]{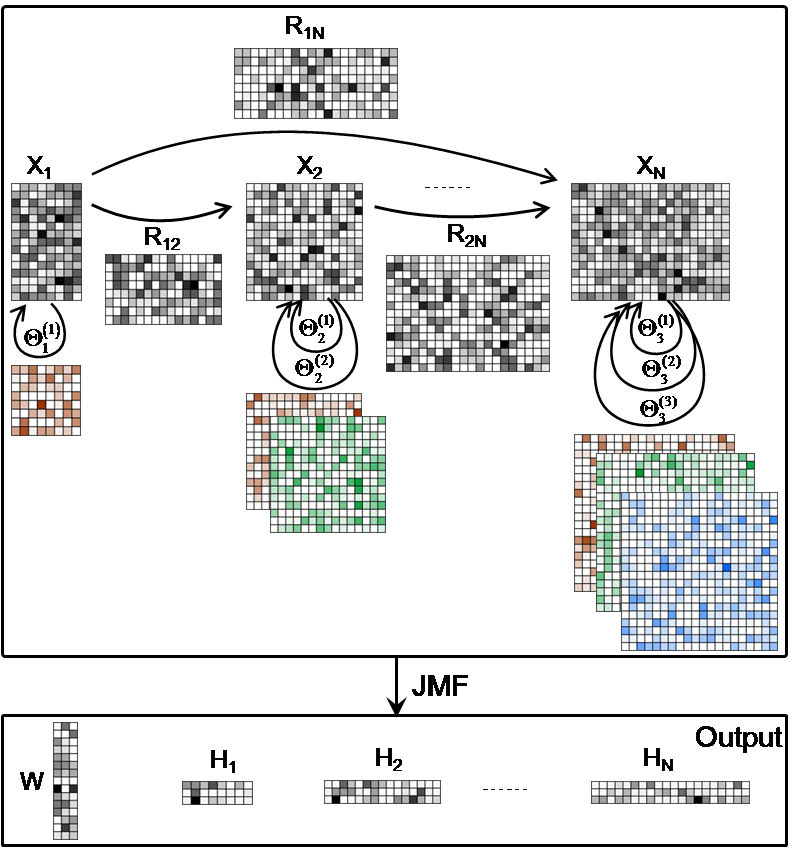}}
\caption{An illustration of JMF. There are $N$ type data matrices $X_{I}$, and many constraints matrices. For example, data matrix $\Theta _{2}^{(1)}$ is the first constraint data matrix on data type $X_{2}$, and $R_{12}$ relates object types $X_{1}$ and $X_{2}$.} \label{Framework}
\end{figure}

Naturally, with the emergence of various kinds of muti-view, within-view and between-view type data, a unified framework is urgently needed. Therefore, we present a generalized form of JMF framework (Figure \ref{Framework}) as follows,
\begin{equation}\label{gernalA}
  \mbox{\small$\displaystyle
  \begin{aligned}
       F(W,{{H}_{1}},\ldots ,{{H}_{N}})= &\sum\limits_{I=1,2,\ldots ,N}{\left\| {{X}_{I}}-W{{H}_{I}} \right\|_{F}^{2}} \\
    & - {{\lambda}_{1}}\sum\limits_{I}{\sum\limits_{t}{T}r({{H}_{I}}\Theta _{I}^{(t)}H_{I}^{T})} \\
                                                       & - {{\lambda }_{2}}\sum\limits_{I\ne J}{Tr({{H}_{I}}{{R}_{IJ}}H_{J}^{T})} + {{\gamma }_{1}}\left\| W \right\|_{F}^{2} \\
                                                       & + {{\gamma }_{2}}\left(\sum\limits_{I}{\sum\limits_{j}{\left\| h_{j}^{I} \right\|_{1}^{2}}}\right), \\
\end{aligned}
$}
\end{equation}
where $h_{j}^{I}$ is the $j$th column of $H_{I}$, $\Theta _{I}^{(t)}$ is the $t$th constraint matrix on the $I$th object, and ${{R}_{IJ}}$ is the relationship matrix between the $I$th and $J$th objects.

\section{Algorithms for JMF}\label{Algorithm}
Similar to the classical NMF problem, the proposed objective function in Eq.\ref{gernalA} is not convex for all variables $W$, $H_I$ ($I = 1, 2,\dots ,N$).
Therefore, it is unrealistic to expect an algorithm to find the global minimum of the proposed optimization problem. For the classical NMF problem, it is convex for one matrix factor when another is fixed. Therefore, we adopt an alternative update strategy for solving JMF. Specifically, fix $H_I$ ($\forall I$), we can obtain $W$ by solving:
\begin{equation}\label{update W}
  \mbox{\small$\displaystyle
  \begin{aligned}
  \underset{W\geq 0}{\text{min}} F(W)=&\sum_{I=1,2,\ldots ,N}\|X_{I}-WH_{I}\|_F^{2} +\gamma_1\|W\|_F^{2}
  \end{aligned}
$}
\end{equation}
Similarly, fix $W$, we can update $H_I, I = 1, 2,\dots , N$ by solving:
\begin{equation}\label{update H}
  \mbox{\small$\displaystyle
  \begin{aligned}
\underset{H_{I, I=1,2,\dots,N}\geq 0}{\text{min}} &F(H_1,H_2,\dots ,H_N)=\sum\limits_{I}{\left\| {{X}_{I}}-W{{H}_{I}} \right\|_{F}^{2}}   \\
    & - {{\lambda}_{1}}\sum\limits_{I}{\sum\limits_{t}{T}r({{H}_{I}}\Theta _{I}^{(t)}H_{I}^{T})} \\
                                                        &- {{\lambda }_{2}}\sum\limits_{I\ne J}{Tr({{H}_{I}}{{R}_{IJ}}H_{J}^{T})} + {{\gamma }_{2}}\left(\sum\limits_{I}{\sum\limits_{j}{\left\| h_{j}^{I} \right\|_{1}^{2}}}\right), \\
\end{aligned}
$}
\end{equation}
We can further update $H_I$ one by one. For any $H_{I}$, given $W$ and $H_{J}, J\ne I $, the objective function for optimizing $H_{I}$ is
\begin{equation}\label{update H_I}
  \mbox{\small$\displaystyle
  \begin{aligned}
\underset{{H}_{I}\ge 0}{\text{min}}
F({{H}_{I}})=&\left\| {{X}_{I}}-W{{H}_{I}} \right\|_{F}^{2}-{{\lambda }_{1}}\sum\limits_{t}{Tr({{H}_{I}}\Theta _{I}^{(t)}H_{I}^{T})} \\
&-{{\lambda }_{2}}\sum\limits_{J\ne I}{Tr({{H}_{I}}{{R}_{IJ}}H_{J}^{T})}+{{\gamma }_{2}}\left(\sum\limits_{j}{\left\| h_{j}^{I} \right\|_{1}^{2}}\right).
\end{aligned}
$}
\end{equation}

Various types of methods have been proposed to solve each subproblem of classical NMF \cite{lee2001algorithms,lin2007projected,kim2008nonnegative,kim2008toward,guan2012nenmf,zhang2014new}. The most widely used approach is the multiplicative update (MUR) algorithm \cite{lee2001algorithms}. This algorithm is easy to implement but converges slowly. And it cannot guarantee the convergence of a local minimum solution. As the resulted matrix factors are nonnegative, Lin treated each subproblem as a bounded constraint optimization problem and used a projected gradient (PG) method to solve it \cite{lin2007projected}. However, PG is inefficient because the Armijo rule is used for searching step size, which is very time-consuming. As the low-rank matrices of the classical NMF are desirable to be sparse, the active set strategy may be a promising method. Kim and Park adopted an active set (AS) method to solve such types of subproblems, which divides variables into an active set and a passive set. In each iteration, AS exchanges only one variable between these two sets \cite{kim2008nonnegative}. They further used the block pivoting strategy to accelerate the AS method (BP) \cite{kim2008toward}. Both AS and BP methods assume that each subproblem is strictly convex, which might bring about numerical instability. As each subprobelm is a convex function and its gradient is Lipchitz continuous, Guan \emph{et al.} solved each subproblem by Nesterov's optimal gradient (Ne) method (NeNMF) \cite{guan2012nenmf}. NeNMF converges faster than previous methods as it has neither time-consuming line search step, nor numerical instability problem. Moreover, NeNMF can be extended to sparse and network regularization even it is not convex. Recently, Zhang \emph{et al.} proposed a new proximal alternating nonnegative least squares (PANLS) to solve each subproblem, which switches between the constrained PG step and unconstrained active set step \cite{zhang2014new}. Luckily, MUR, PG, Ne and PANLS are all suitable for solving JMF, while both AS and BP are not directly applicable to the network-regularized NMF. As noted that the current code of BP needs to be modified and it may not be efficient if the BP update method is used \cite{zhang2014new}. In the following subsections, we develop four update methods (MUR, PG, Ne and PANLS) for optimizing JMF in spirit of the above exploration, and present their corresponding algorithms in Appendix Algorithms 1-4, respectively.


\subsection{Multiplicative update algorithm}
Firstly, we solve JMF with the MUR algorithm which searches along a rescaled gradient direction with a fixed form of learning rate to guarantee the nonnegativity of the low-rank matrices. The details of MUR are shown as follows. The Lagrange function $L$ is
$L(W,{{H}_{I}})=F+Tr(\Psi {{W}^{T}})+\sum_{I}{Tr({{\Phi }_{I}}H_{I}^{T})}$,
$\Psi =\left[ {{\psi }_{ij}} \right]$ and $\Phi_{I} =\left[ \phi _{ij}^{I} \right]$. The partial derivative of $L$ with respect to $W$ and $H_{I}$ are respectively as follows:
\begin{equation}
  \mbox{\small$\displaystyle
\left\{ \begin{aligned}
  \frac{\partial L}{\partial W}=&\sum\limits_{I}{[-2{{X}_{I}}H_{I}^{T}+2W{{H}_{I}}H_{I}^{T}]+2{{\gamma }_{1}}W+\Psi }, \\
  \frac{\partial L}{\partial {{H}_{I}}}=&-2{{W}^{T}}{{X}_{I}}+2{{W}^{T}}W{{H}_{I}}-{{\lambda }_{1}}\sum\limits_{t}{{{H}_{I}}}[\Theta _{I}^{(t)}+{{(\Theta _{I}^{(t)})}^{T}}] \\
  &-{{\lambda }_{2}}\sum\limits_{J\ne I}{{{H}_{J}}R_{IJ}^{T}} + {{\gamma }_{2}}2{{e}_{K\times K}}{{H}_{I}}+{{\Phi }_{I}}, (I = 1,\dots ,N).
\end{aligned} \right.
$}
\end{equation}
Based on the KKT conditions ${{\psi }_{ij}}{{W}_{ij}}=0\text{ and }\phi _{ij}^{I}{{({{H}_{I}})}_{ij}}=0$, we get the following equations for ${{W}_{ij}},{{({{H}_{I}})}_{ij}}$, respectively,
{\footnotesize $$\left\{ \begin{aligned}
 &-2\sum\limits_{I}{{{({{X}_{I}}H_{I}^{T})}_{ij}}{{W}_{ij}}+\left[2\sum\limits_{I}{(}W{{H}_{I}}H_{I}^{T})+2{{\gamma }_{1}}W \right]_{ij}}{{W}_{ij}}=0, \\
 & {{\left[ -{{W}^{T}}{{X}_{I}}-\frac{{{\lambda }_{1}}}{2}\sum\limits_{t}{{{H}_{I}}}(\Theta _{I}^{(t)}+{{(\Theta _{I}^{(t)})}^{T}})-\frac{{{\lambda }_{2}}}{2}\sum\limits_{J\ne I}{H_{J}^{T}R_{IJ}^{T}} \right]}_{ij}} \\
& \times{{({{H}_{I}})}_{ij}}+{{({{W}^{T}}W{{H}_{I}}+{{\gamma }_{2}}{{e}_{K\times K}}{{H}_{I}})}_{ij}}{{({{H}_{I}})}_{ij}}=0, (I = 1,\dots ,N). \\
\end{aligned} \right.$$}
Then we can get the following update rules:
{\small\begin{equation}\label{update}
  \begin{cases}
     &{{w}_{ij}}\leftarrow {{w}_{ij}}\frac{{{\left( \sum\limits_{I}{{{X}_{I}}H_{I}^{T}} \right)}_{ij}}}{{{\left( \sum\limits_{I}{W{{H}_{I}}H_{I}^{T}}+{{\gamma }_{1}}W \right)}_{ij}}},  \\
   &h_{ij}^{I}\leftarrow h_{ij}^{I}\frac{{{\left( {{W}^{T}}{{X}_{I}}+\frac{{{\lambda }_{1}}}{2}\sum\limits_{t}{{{H}_{I}}[}{{\Theta }_{I}}+{{(\Theta _{I}^{(t)})}^{T}}]+\frac{{{\lambda }_{2}}}{2}\sum\limits_{J\ne I}{{{H}_{J}}}R_{IJ}^{T} \right)}_{ij}}}{{{\left( ({{W}^{T}}W+{{\gamma }_{2}}{{e}_{K\times K}}){{H}_{I}} \right)}_{ij}}},  \\
  \end{cases}
\end{equation}}

Note that the usual stopping criterion for MUR is
 \begin{equation}\label{stop 1}
  \mbox{\small$\displaystyle
  \begin{aligned}
 \frac{F({{W}^{t}},H_{1}^{t},\ldots ,H_{I}^{t})-F({{W}^{t+1}},H_{1}^{t+1},\ldots ,H_{I}^{t+1})}{F({{W}^{1}},H_{1}^{1},\ldots ,H_{I}^{1})-F({{W}^{t\text{+}1}},H_{1}^{t+1},\ldots ,H_{I}^{t+1})}\le \tau,
 \end{aligned}
$}
\end{equation}
where $\tau $ is a predefined tolerance. While the usual stopping criterion used in the other three update methods is
 \begin{equation}\label{stop 2}
  \mbox{\footnotesize$\displaystyle
  \begin{aligned}
  & {{\left\| [\begin{aligned}
   {{\nabla }_{W}}F{{({{W}^{t}},H_{1}^{t},\ldots ,H_{N}^{t})}^{T}} \\
\end{aligned},\ldots, {{\nabla }_{{{H}_{N}}}}F({{W}^{t}},H_{1}^{t},\ldots ,H_{N}^{t})] \right\|}_{F}} \\
 & \le \tau {{\left\| [\begin{aligned}
   {{\nabla }_{W}}F{{({{W}^{1}},H_{1}^{1},\ldots ,H_{N}^{1})}^{T}},\ldots , & {{\nabla }_{{{H}_{N}}}}F({{W}^{1}},H_{1}^{1},\ldots ,H_{N}^{1})  \\
\end{aligned}] \right\|}_{F}}. \\
\end{aligned}
$}
\end{equation}
{\small If $\left[ {{\nabla }_{W}}F({{W}^{t}},H_{1}^{t},\cdots ,H_{N}^{t}),\cdots ,{{\nabla }_{{{H}_{N}}}}F({{W}^{t}},H_{1}^{t},\cdots ,H_{{{H}_{N}}}^{t}) \right]$} is denoted by ${{\nabla }^{t}}$, then Eq. \ref{stop 2} can  be represented by ${{\left\| {{\nabla }^{t}} \right\|}_{F}}\le \tau {{\left\| {{\nabla }^{1}} \right\|}_{F}}$. Note that ${\nabla}^{t}$ may vary slowly when the variables close to a stationary point. Thus, we terminate PG, Ne and PANLS, when Eq. \ref{stop 2} or the following Eq. \ref{stop 22} is satisfied,
\begin{equation}\label{stop 22}
\left| {{\left\| {{\nabla }^{t+10}} \right\|}_{F}}-{{\left\| {{\nabla }^{t}} \right\|}_{F}} \right|\le {{10}^{-3}}\tau {{\left\| {{\nabla }^{1}} \right\|}_{F}}. \end{equation}

Generally, MUR is simple and easy to implement, and it quickly decreases the objective value at the beginning. But it does not guarantee the convergence to any local minimum because its solution is unnecessarily a stationary point. Even though it has a stationary point, it converges slowly. If some rows or columns of $X_{I}$ are close to zero, the result may have numerical problems.

\subsection{Projected gradient algorithm}
We adopt PG to solve each subproblem, which uses the Armijo rule to search the step size along the projection arc. We take the subproblem Eq. \ref{update W} as an example. The step size $\alpha$ satisfies:
\begin{equation}
  \mbox{\footnotesize$\displaystyle
  \begin{aligned}
(1-\sigma)\left\langle \nabla F(\overset{-}{\mathop{W}}\,) \right.,\left. \overset{\sim }{\mathop{W}}\,-\overset{-}{\mathop{W}}\, \right\rangle +\frac{1}{2}\left\langle \overset{\sim }{\mathop{W}}\,-\overset{-}{\mathop{W}}\,, \right.\left. Q(\overset{\sim }{\mathop{W}}\,-\overset{-}{\mathop{W}}\,) \right\rangle \le 0,
\end{aligned}
$}
\end{equation}
where $\sigma\in (0.1)$ ($\sigma$=0.01 is used), {\small$\overset{\sim }{\mathop{W}}\,\equiv P[\overset{-}{\mathop{W}}\,-\alpha \nabla F(\overset{-}{\mathop{W}}\,)]$} and $Q$ is the second moment matrix of $F(W)$. The gradient function of $W$ is
$$ \nabla F(W)=2\sum\limits_{I}{(W{{H}_{I}}}H_{I}^{T}-{{X}_{I}}H_{I}^{T})+2{{\gamma }_{1}}W. $$
The Hessian matrix for $W$ is
$$ {Q}_{W}=2\sum\limits_{I}{{{H}_{I}}H_{I}^{T}\otimes {{I}_{M}}}+2{{\gamma }_{1}}{{I}_{K}}\otimes {{I}_{M}}, $$
where $\otimes $ is Kronecker product. The PG is very easy to implement but it is time-consuming and may suffer from the zigzag phenomenon when approaching the local minimizer if the condition number is bad.

\subsection{Nesterov's optimal gradient algorithm}
NetNMF updates two sequences recursively to optimize each low-rank matrix. One sequence stores the approximate solutions which are obtained by PG method on the search points with step size determined by the Lipchitz constant. Another sequence stores the search points which are the combination of the latest two approximation solution. In this way, the objective function is convex for the variable and the gradient of the objective function is Lipchitz continuous that are the two prerequisites when applying Nesterov's method \cite{nesterov1983method,nesterov2013introductory,nesterov2007gradient}. NeNMF can be conveniently extended for optimizing $L_1$-norm, $L_2$-norm and network-regularized NMF and can also been extended for JMF. Given ${{H}_{I}}$ ($I=1,2,\ldots ,N$), the objective function $F(W)$ for optimizing $W$ in Eq. \ref{update W}
is a convex function, and the gradient function for $W$ satisfies Lipschitz continuity as follows,
\begin{equation}
\mbox{\small$\displaystyle
\begin{aligned}
    &{{\left\| \nabla F({{W}_{1}})-\nabla F({{W}_{2}}) \right\|}_{F}}  \\
=   & 2{{\left\| ({{W}_{1}}-{{W}_{2}})\left(\sum\limits_{I}{{{H}_{I}}H_{I}^{T}}+{{\gamma }_{1}}{{I}_{K}}\right) \right\|}_{F}} \\
\le & 2{{\left\| \sum\limits_{I}{{{H}_{I}}H_{I}^{T}+{{\gamma }_{1}}{{I}_{K}}} \right\|}_{2}}\times {{\left\| {{W}_{1}}-{{W}_{2}} \right\|}_{F}}, \\
\end{aligned}
$}
\end{equation}
where $\nabla F(W)=2\sum\limits_{I}{(W{{H}_{I}}}H_{I}^{T}-{{X}_{I}}H_{I}^{T})+2{{\gamma }_{1}}W$ is the gradient of $F(W)$.
Though the objective function $F(H_{I})$ in Eq. \ref{update H_I} for optimizing $H_{I}$ is nonconvex,
the gradient function for $H_{I}$ satisfies Lipschitz continuity as follows,
\begin{equation}
\mbox{\small$\displaystyle
  \begin{aligned}
  & {{\left\| \nabla F({{H}_{I}^1})-\nabla F({{H}_{I}^2}) \right\|}_{F}} \\
  =&{{\left\| 2({{W}^{T}}W+{{\gamma }_{2}}{{e}_{K}})({{H}_{I}^1}-{{H}_{I}^2})-{{\lambda }_{1}}({{H}_{I}^1}-{{H}_{I}^2})({{\Theta }_{I}}+\Theta _{I}^{T}) \right\|}_{F}} \\
  \le &\left(2{{\left\| {{W}^{T}}W+{{\gamma }_{2}}{{e}_{K}} \right\|}_{2}} + {{\lambda }_{1}}{{\left\| \sum\limits_{t}{\Theta _{I}^{(t)}+{{(\Theta _{I}^{(t)})}^{T}}} \right\|}_{2}}\right) \\
      &\times {{\left\| {{H}_{I}^1}-{{H}_{I}^2} \right\|}_{F}}, \\
\end{aligned}
$}
\end{equation}
where {\small$\nabla F({{H}_{I}})=-2{{W}^{T}}{{X}_{I}}+2{{W}^{T}}W{{H}_{I}}-{{\lambda }_{1}}\sum\limits_{t}{{{H}_{I}}}[\Theta _{I}^{(t)}+{{(\Theta _{I}^{(t)})}^{T}}]-{{\lambda }_{2}}\sum\limits_{J\ne I}{{{H}_{J}}R_{IJ}^{T}}+{{\gamma }_{2}}2{{e}_{K}}{{H}_{I}}$}
is the gradient of $F(H_{I})$. Ne indeed decrease the objective function but cannot guarantee the convergence to any stationary point as the objection function $F(H_{I})$ is nonconvex.

\subsection{Proximal alternating nonegative least squares algorithm}
Inspired by the PANLS for solving the typical NMF problem \cite{zhang2014new}, we adopt the kernel ideal for solving JMF. The subproblems can be transformed as follows,
{\small\begin{equation}\label{update W PANLS}
\begin{aligned}
{{W}^{k+1}}=\underset{W\ge 0}{arg\,\text{min}}
   &\sum\limits_{I=1}^{N}{\left\| {{X}_{I}}-WH_{I}^{k} \right\|_{F}^{2}\text{+}{{\gamma }_{1}}\left\| W \right\|_{F}^{2}} \\
   &+{{\tau }_{1}}\left\| W-{{W}^{k}} \right\|_{F}^{2},
   \end{aligned}
   \end{equation}}

{\small\begin{equation}\label{update H_I PANLS}
\begin{aligned}
H_{I}^{k+1}= &\underset{{H}_{I}\ge 0}{arg\,\text{min}}
  \left\| {{X}_{I}}-{{W}^{k}}{{H}_{I}} \right\|_{F}^{2}-{{\lambda }_{1}}\sum\limits_{t}{Tr({{H}_{I}}\Theta _{I}^{(t)}H_{I}^{T})} \\
  &-{{\lambda }_{2}}\sum\limits_{J\ne I}{Tr({{H}_{I}}{{R}_{IJ}}H_{J}^{T})}
  \text{+}{{\gamma }_{2}}\left(\sum\limits_{j}{\left\| h_{j}^{I} \right\|_{1}^{2}}\right) \\
  &+\tau _{2}^{I}\left\| {{H}_{I}}-H_{I}^{k} \right\|_{F}^{2}.
  \end{aligned}
  \end{equation}}

The Hessian matrices of $F(W)$ and $F(H)$ are:
\begin{equation}
  \mbox{\small$\displaystyle
  \begin{aligned}
 Q(W)=2\left(\sum\limits_{I}{{H}_{I}}H_{I}^{T}\right)\otimes {{I}_{M}}+2({{\gamma }_{1}}+{{\tau }_{1}}){{I}_{K}}\otimes {{I}_{M}},
\end{aligned}
$}
\end{equation}
\begin{equation}
  \mbox{\small$\displaystyle
  \begin{aligned}
Q({{H}_{I}})=& 2{{I}_{{{N}_{I}}}}\otimes ({{W}^{T}}W)-{{\lambda }_{1}} \left(\sum\limits_{t}({{\Theta }^{(t)}})^{T}+\sum\limits_{t}{{{\Theta }^{(t)}}}\right)\otimes {{I}_{K}} \\
  &+ 2({{\gamma }_{2}}+\tau _{2}^{I}){{I}_{{{N}_{I}}}}\otimes {{I}_{K}}.
\end{aligned}
$}
\end{equation}
Thus, $F(W)$ and $F(H_{I})$ are strictly convex function of variables $W$ and $H_{I}$ with proper $\tau_{1}$ and $\tau _{2}^{I}$. And each subproblem has a unique minimizer according to Frank Wolfe theorem. Therefore, PANLS has a nice convergence property.

\section{Prediction models based on JMF}\label{Prediction}
NMF and its variants can be used for prediction tasks \cite{stravzar2016orthogonal}. JMF can also be extended to the prediction form. Both the basis matrix and coefficients matrices can be used for prediction. The prediction based on the basis matrix is denoted by JMF/L, while the prediction based on the coefficient matrices is denoted by JMF/R. Let $X_I$ ($I = 1, 2,\dots ,N$) be the training datasets and $\overset{\sim}{X_{I}}$ ($I = 1, 2,\dots, N$) be the testing datasets. We can obtain low-rank matrices $W$ and $H_I$ $(I = 1, 2,\dots ,N)$ by JMF on the training datasets.

For JMF/L, fix the learned coefficients matrices $H_I$ ($I = 1, 2,\dots ,N$), the predicted factor $W$ can be obtained by solving Eq. \ref{update W} on the testing datasets. There were two prediction scenarios based on the learned basis matrix on the training data. In the scenario I for class prediction of new samples, the basis matrix is used as the prediction factor, which can be obtained based on the testing data and learned coefficient matrices $H_I$. The prediction class of each sample can be obtained based on the maximum value in each row of $\widehat{W}$. In the scenario II for one view data prediction (e.g., $X_1$) from other view data ($X_I$, $I=2,...,N$), the new basis matrix $\widehat{W}$ is computed with the learned coefficient matrices and the testing data $X_{2}$, \dots, $X_{N}$. Then the multiplication of $\widehat{W}$ and the learned coefficient matrix $H_{1}$ is used to predict $\widehat{X_{1}}$. In this paper, we illustrated the second scenario of JMF/L with one real application.

Similarly, for JMF/R, fix the learned basis matrix $W$, the predicted factors $H_I$ $(I = 1, 2,\dots ,N)$ can be obtained by solving Eq. \ref{update H_I}, which can be used as prediction factors as the scenario I in JMF/L.

\begin{figure*}[htp]
\centerline{\includegraphics[width=0.9\textwidth]{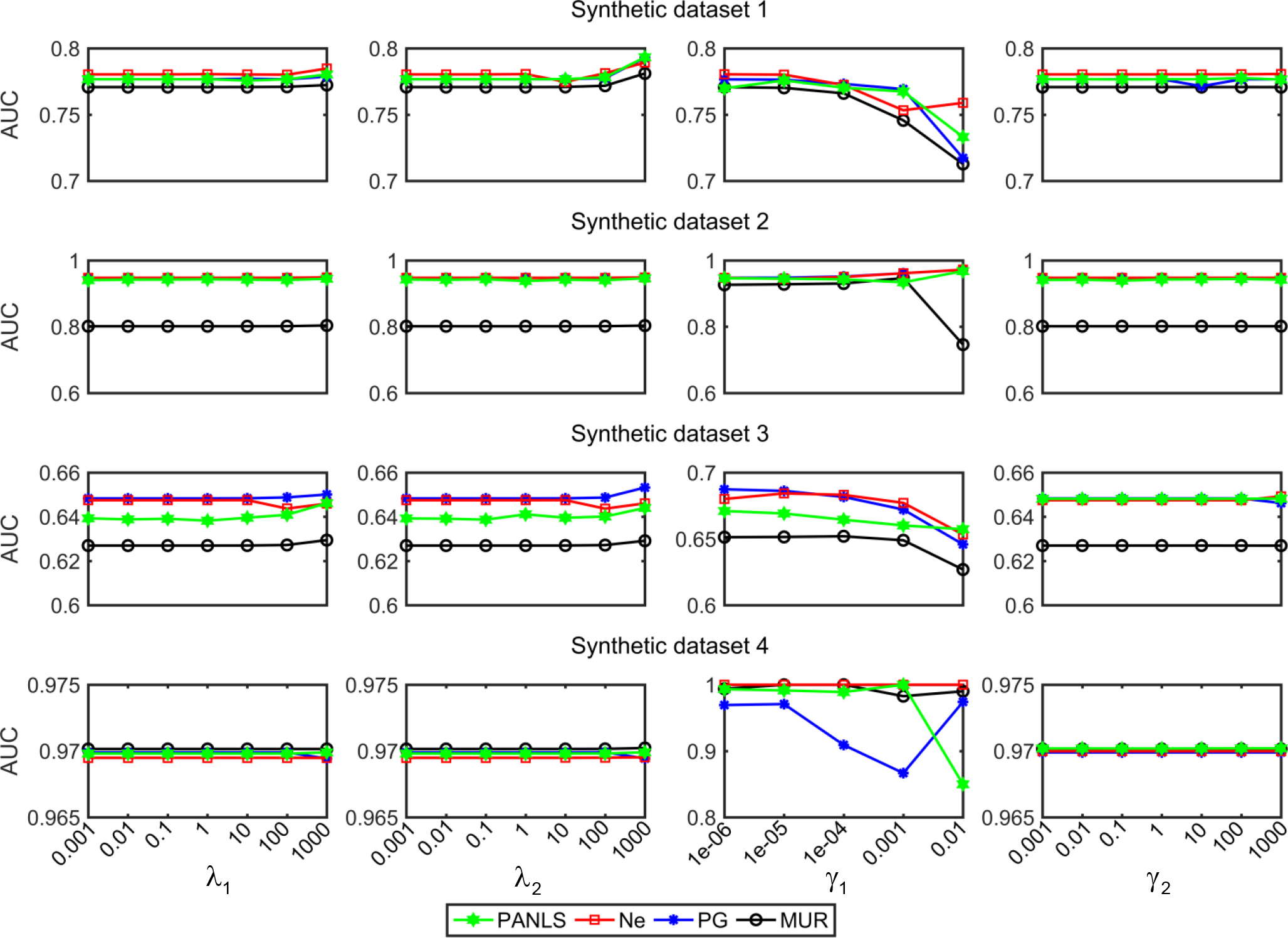}}
\caption{The influence of each constraint term on the performance of JMF.} \label{Constraints influence}
\end{figure*}

\section{Experiments}\label{Experiments}
To demonstrate the performance of JMF, we applied it to four synthetic and three real datasets. Firstly, we evaluated how the parameters influence the performance of JMF in terms of the Area Under the Curve (AUC) on four synthetic datasets. Then we compared the average objective values with respect to iteration numbers or running time of the four update methods on the synthetic datasets. Finally, we applied JMF to three real data from diverse fields. We run the experiments of synthetic datasets on a machine with Intel Core i7-4770 CPU @ 3.40GHz ¡Á4 with 16 GB RAM and used MATLAB (R2016a) 64-bit for the general implementation. The real datasets were run on a windows server with Intel (R) Xeon (R) E5-2643 v3 CPU @ 3.40GHz ¡Á2 with 768 GB RAM and implemented on MATLAB (R2013a) 64-bit. For the purpose of reproducibility, the data and code are available at: http://page.amss.ac.cn/shihua.zhang/software.html

\subsection{Synthetic dataset 1}\label{RThe}
We adopt a similar simulation strategy as used in Experiment $1$ in \cite{wu2016stability} to demonstrate the effectiveness of the algorithms for JMF. The true low-rank $=4$ and the ground truth basis matrix represented by ${{W}_{0}}\in R_{+}^{45\times 4}$ was constructed with \emph{coph} = 0 as follows,
\begin{equation}
  \mbox{$\displaystyle
  {{W}_{0}}[j,k]=\left\{ \begin{aligned}
  & 1,    1+x_{k}(10) \le j \le 10+x_{k}(10), \\
  & 0,    \mbox{otherwise}. \\
 \end{aligned} \right.
$}
\end{equation}
Meanwhile, three coefficient matrices (${{H}_{01}}\in R_{+}^{4\times 130},{{H}_{02}}\in R_{+}^{4\times 170},{{H}_{03}}\in R_{+}^{4\times 215}$) were constructed with \emph{coph} =0,
\begin{equation}
  \mbox{$\displaystyle
{{H}_{01}}[j,k]=\left\{ \begin{aligned}
  & 1,  1+x_{j}(30) \le k \le 30+x_{j}(30), \\
  & 0,\mbox{otherwise}. \\
\end{aligned} \right.
$}
\end{equation}
\begin{equation}
  \mbox{ $\displaystyle
{{H}_{02}}[j,k]=\left\{ \begin{aligned}
  & 1,  1+x_{j}(40) \le k\le 40+x_{j}(40), \, j \ne 4,\\
  & 0, \mbox{otherwise}.  \\
\end{aligned} \right.
$}
\end{equation}
\begin{equation}
  \mbox{ $\displaystyle
{{H}_{03}}[j,k]=\left\{ \begin{aligned}
 & 1,    1+x_{j}(50) \le k\le 50+x_{j}(50), \, j\ne 3, \\
 & 0,\text{    otherwise}\text{.} \\
\end{aligned} \right.
$}
\end{equation}
where $x_{j}(n)=(j-1)(n-\textit{coph})$.

We set the data matrices by ${{X}_{0I}}={{W}_{0}}{{H}_{0I}}+\mu E$ ($I = 1, 2, 3$), where $E$ was Gaussian noise and $\mu=2$. The within constraint on each data matrix was simulated as follows,
\begin{equation}
  \mbox{\small$\displaystyle
\Theta _{I}^{k}[s,t]=\left\{ \begin{aligned}
  & 1,    \mbox{if}\, {{H}_{I}}[k,s]=1 \,\mbox{and}\, {{H}_{I}}[k,t]=1, \\
 & 0,    \mbox{otherwise}. \\
\end{aligned} \right.
$}
\end{equation}
We obtained the within constraint matrix on the $I$th source by averaging the value of $\Theta_{I}$ and ${\Theta_{I}}^{T}$, where ${{\Theta }_{I}}=\sum\limits_{k}{\Theta _{I}^{k}} + 0.1E, (I=1,2,3)$. The between constraint $R_{IJ}$ on the $I$th and $J$th data matrices was simulated as follows,
\begin{equation}
  \mbox{\small$\displaystyle
R_{IJ}^{k}[s,t]=\left\{ \begin{aligned}
  & 1,\mbox{if}\, {{H}_{I}}[k,s]=1 \,\mbox{and}\, {{H}_{J}}[k,t]=1, \\
 & 0,\mbox{otherwise}. \\
\end{aligned} \right.
$}
\end{equation}
The between constraint matrix ${{R}_{IJ}}=\sum\limits_{k}{R_{IJ}^{k}} + 0.1E$.

\subsection{Synthetic dataset 2}
We simulated a relative large-scale dataset with the true low rank $=10$. Different from dataset 1, the entries of the true basis matrix ${{W}_{0}}\in R_{+}^{1000\times 10}$ were deemed as independent and identical Bernoulli variables with probability equals to $1/10$. And we constructed the ground truth coefficient matrices ${{H}_{01}}\in R_{+}^{10\times 200},{{H}_{02}}\in R_{+}^{10\times 300},{{H}_{03}}\in R_{+}^{10\times 500}$ in the following manner with \emph{coph}=$0$, $5$, $10$, respectively.
$${{H}_{01}}[j,k]=\left\{ \begin{aligned}
  & 1,   1+x_{j}(20)\le k\le 20+x_{j}(20), \\
  & 0,   \mbox{otherwise}. \\
\end{aligned} \right.$$
$${{H}_{02}}[j,k]=\left\{ \begin{aligned}
  & 1,    1+x_{j}(30)\le k\le 30+x_{j}(30), \\
  & 0,    \mbox{otherwise}. \\
\end{aligned} \right.$$
$${{H}_{03}}[j,k]=\left\{ \begin{aligned}
  & 1,    1+x_{j}(50)\le k\le 50+x_{j}(50), \\
  & 0,    \mbox{otherwise}. \\
\end{aligned} \right.$$
We set data matrices by ${{X}_{0i}}={{W}_{0}}{{H}_{0i}}+\mu E$, ($I = 1, 2, 3$), where $E$ was Gaussian noise and $\mu =2$. Similarly, ${{\Theta }_{i}}$ and ${{R}_{ij}}$ were generated as mentioned above.

\begin{table*}[htbp]
  \centering
  \begin{threeparttable}[b]
  \caption{Performance comparison of update algorithms on synthetic dataset 1.}
  \label{tabsolverComparison1}
    \begin{tabular}{l|rrr|rrr|rrr|rrr}
    \hline
    \multirow{2}[2]{*}{} & \multicolumn{3}{c|}{Time (seconds)} & \multicolumn{3}{c|}{\#iteration} & \multicolumn{3}{c|}{Reconstruction error} & \multicolumn{3}{c}{AUC} \\  \hline
    Stop1 & $10^{-5}$ & $10^{-6}$ & $10^{-7}$ & $10^{-5}$ & $10^{-6}$ & $10^{-7}$ & $10^{-5}$ & $10^{-6}$ & $10^{-7}$ & $10^{-5}$ & $10^{-6}$ & $10^{-7}$ \\
    \hline
    MUR   & \textbf{2.96} & 5.68  & 10.03  & 341   & 658   & 1156  & 31042.90  & 31026.51  & 31022.75  & 0.76  & 0.77  & 0.77  \\
    PG    & 20.78  & 25.18  & 30.53  & 76    & 93    & 114   & \textbf{31014.42} & \textbf{31013.91} & \textbf{31013.84} & \textbf{0.78} & \textbf{0.78} & \textbf{0.78} \\
    Ne    & 9.72  & 12.15  & 15.01  & 68    & 85    & 106   & 31014.43  & \textbf{31013.91}  & \textbf{31013.84} & \textbf{0.78} & \textbf{0.78} & \textbf{0.78} \\
    PANLS & 3.28  & \textbf{4.04} & \textbf{5.03} & 78    & 97    & 125   & 31015.16  & 31013.95  & 31013.85  & \textbf{0.78} & \textbf{0.78} & \textbf{0.78} \\
    \hline
    Stop1+ &       &       & \multicolumn{1}{r}{} &       &       & \multicolumn{1}{r}{} &       &       & \multicolumn{1}{r}{} &       &       &  \\
    \hline
    MUR   & \textbf{2.31} & 5.08  & 6.04  & 339   & 729   & 873   & 31416.28  & 31398.34  & 31394.33  & 0.78  & 0.79  & 0.79  \\
    PG    & (2) 9.01 & (2) 13.27 & (2) 21.37 & 34    & 51    & 82    & 31380.58  & 31359.02  & 31348.51  & 0.79  & 0.80  & \textbf{0.81} \\
    Ne    & 7.85  & 11.59  & 16.33  & 56    & 82    & 118   & \textbf{31331.53} & \textbf{31318.13} & \textbf{31311.99} & 0.79  & 0.79  & 0.80  \\
    PANLS & \textbf{(3) 3.68} & \textbf{(3) 4.60} & \textbf{(3) 5.56} & 54    & 68    & 84    & 31509.66  & 31507.87  & 31508.37  & \textbf{0.81} & \textbf{0.81} & \textbf{0.81} \\
    \hline
    \multicolumn{1}{r}{} &       &       & \multicolumn{1}{r}{} &       &       & \multicolumn{1}{r}{} &       &       & \multicolumn{1}{r}{} &       &       &  \\
    \hline
    Stop2 & $10^{-2}$ & $10^{-3}$ & $10^{-4}$ & $10^{-2}$ & $10^{-3}$ & $10^{-4}$ & $10^{-2}$ & $10^{-3}$ & $10^{-4}$ & $10^{-2}$ & $10^{-3}$ & $10^{-4}$ \\
    \hline
    PG    & 39.22  & 39.22  & 39.22  & 146   & 146   & 146   & \textbf{31013.84} & \textbf{31013.84} & \textbf{31013.84} & \textbf{0.78} & \textbf{0.78} & \textbf{0.78} \\
    Ne    & 19.36  & 19.36  & 19.36  & 137   & 137   & 137   & \textbf{31013.84} & \textbf{31013.84} & \textbf{31013.84} & \textbf{0.78} & \textbf{0.78} & \textbf{0.78} \\
    PANLS & \textbf{0.63} & \textbf{2.05} & \textbf{4.19} & 13    & 51    & 105   & 31262.01  & 31027.84  & 31014.02  & 0.71  & \textbf{0.78} & \textbf{0.78} \\
    \hline
    Stop2+ &       &       &       &       &       &       &       &       &       &       &       &  \\
    \hline
    PG    & (1) 32.86 & (1) 32.86 & (1) 32.86 & 125   & 125   & 125   & 31346.22  & 31346.22  & 31346.22  & \textbf{0.81} & \textbf{0.81} & \textbf{0.81} \\
    Ne    & 21.73  & 21.73  & 21.73  & 153   & 153   & 153   & \textbf{31307.93} & \textbf{31307.93} & \textbf{31307.93} & 0.80  & 0.80  & 0.80  \\
    PANLS & \textbf{(1) 8.02} & \textbf{(1) 8.02} & \textbf{(1) 8.02} & 120   & 120   & 120   & 31500.84  & 31500.84  & 31500.84  & \textbf{0.81} & \textbf{0.81} & \textbf{0.81} \\
     \hline
    \end{tabular}
    \begin{tablenotes}
    \item \small{The number in the bracket represents the frequency of the algorithm's iteration exceeds $2000$ under the $10$ initial values. The Stop1 and Stop2 represent the algorithms terminate with the first and second stop criteria respectively and the regularization being empty, while Stop1+ and Stop2+ represent the algorithms terminate with the first and second stop criteria respectively and the regularization being nonempty. }
    \end{tablenotes}
    \end{threeparttable}
  \label{tab:addlabel}%
\end{table*}%

\begin{table*}[htbp]
  \centering
  \begin{threeparttable}[b]
  \caption{Performance comparison of update algorithms on synthetic dataset 2.}
  \label{tabsolverComparison2}
    \begin{tabular}{l|rrr|rrr|rrr|rrr}
    \hline
    \multirow{2}[2]{*}{} & \multicolumn{3}{c|}{Time (seconds)} & \multicolumn{3}{c|}{\#iteration} & \multicolumn{3}{c|}{Reconstruction error} & \multicolumn{3}{c}{AUC} \\ \hline
    Stop1 & $10^{-5}$ & $10^{-6}$ & $10^{-7}$ & $10^{-5}$ & $10^{-6}$ & $10^{-7}$ & $10^{-5}$ & $10^{-6}$ & $10^{-7}$ & $10^{-5}$ & $10^{-6}$ & $10^{-7}$ \\
    \hline
    MUR   & \textbf{(2) 9.68} & (2) 26.89 & (2) 84.20 & 199   & 558   & 1724  & 1425015.76  & 1421570.53  & 1420184.96  & 0.89  & 0.93  & 0.94  \\
    PG    & 50.75  & 80.81  & 97.36  & 48    & 74    & 89    & \textbf{1420060.68} & \textbf{1419679.13} & 1419662.75  & \textbf{0.94} & \textbf{0.95} & \textbf{0.95} \\
    Ne    & 37.32  & 58.09  & 70.04  & 47    & 73    & 88    & 1420070.63  & 1419679.79  & \textbf{1419662.69} & \textbf{0.94} & \textbf{0.95} & \textbf{0.95} \\
    PANLS & 13.90  & \textbf{21.74} & \textbf{26.48} & 61    & 100   & 128   & 1420389.53  & 1419711.43  & 1419664.65  & \textbf{0.94} & \textbf{0.95} & \textbf{0.95} \\
    \hline
    Stop1+ &       &       & \multicolumn{1}{r}{} &       &       & \multicolumn{1}{r}{} &       &       & \multicolumn{1}{r}{} &       &       &  \\
    \hline
    MUR   & \textbf{(5) 8.65} & (5) 26.5 & \multicolumn{1}{r}{(5) 62.73} & 197   & 609   & \multicolumn{1}{r}{1443} & 1425298.99  & 1421669.47  & \multicolumn{1}{r}{1420916.36} & 0.90  & 0.95  & 0.95  \\
    PG    & 47.78  & 66.23  & \multicolumn{1}{r}{73.03} & 60    & 82    & \multicolumn{1}{r}{92} & 1424383.52  & 1424099.74  & \multicolumn{1}{r}{1424098.05} & \textbf{0.96} & \textbf{0.97} & \textbf{0.97} \\
    Ne    & (3) 38.19 & (3) 49.81 & \multicolumn{1}{r}{(3) 79.48} & 45    & 59    & \multicolumn{1}{r}{101} & \textbf{1420908.49} & 1420554.45 & \multicolumn{1}{r}{1420491.00} & 0.95  & 0.96  & 0.96  \\
    PANLS & 11.48  & \textbf{21.35} & \multicolumn{1}{r}{\textbf{25.93}} & 46    & 81    & \multicolumn{1}{r}{101} & 1421211.47  & \textbf{1419969.28}  & \multicolumn{1}{r}{\textbf{1419935.96}} & 0.93  & 0.95  & 0.95  \\
    \hline
    \multicolumn{1}{r}{} &       &       & \multicolumn{1}{r}{} &       &       & \multicolumn{1}{r}{} &       &       & \multicolumn{1}{r}{} &       &       &  \\
    \hline
    Stop2 & $10^{-2}$ & $10^{-3}$ & $10^{-4}$ & $10^{-2}$ & $10^{-3}$ & $10^{-4}$ & $10^{-2}$ & $10^{-3}$ & $10^{-4}$ & $10^{-2}$ & $10^{-3}$ & $10^{-4}$ \\
    \hline
    PG    & 9.01  & 119.10  & 119.10  & 11    & 108   & 108   & 1429843.58  & 1419660.98 & 1419660.98 & 0.81  & \textbf{0.95} & \textbf{0.95} \\
    Ne    & 7.54  & 86.41  & 86.41  & 11    & 107   & 107   & \textbf{1427159.80} & \textbf{1419660.97}  & \textbf{1419660.97}  & \textbf{0.85} & \textbf{0.95} & \textbf{0.95} \\
    PANLS & \textbf{2.73} & \textbf{2.78} & \textbf{10.48} & 11    & 11    & 48    & 1429249.16  & 1428993.74  & 1421766.35  & 0.82  & 0.82  & 0.92  \\
    \hline
    Stop2+ &       &       &       &       &       &       &       &       &       &       &       &  \\
    \hline
    PG    & 84.14  & 84.14  & 84.14  & 103   & 103   & 103   & \textbf{1424098.23} & 1424098.23  & 1424098.23  & \textbf{0.97} & \textbf{0.97} & \textbf{0.97} \\
    Ne    & 7.46  & 80.94  & 80.94  & 11    & 110   & 110   & 1427301.13  & \textbf{1420477.00} & \textbf{1420477.00} & 0.86  & 0.96  & 0.96  \\
    PANLS & \textbf{3.01} & \textbf{3.06} & \textbf{33.48} & 11    & 11    & 120   & 1429044.28  & 1428805.69  & 1419755.46  & 0.83  & 0.83  & 0.95  \\
   \hline
    \end{tabular}
    \end{threeparttable}
  \label{tab:addlabel}%
\end{table*}%

\subsection{Synthetic dataset 3}
We simulated a dataset with overlap information on coefficient matrices as well as large noise in prior networks. We set the true low-rank $=20$ and constructed the ground truth basis matrix ${{W}_{0}}\in R_{+}^{2000\times 20}$ with \emph{coph} =$15$,
$${{W}_{0}}[j,k]=\left\{ \begin{aligned}
  & 1,    1+x_{k}(100)\le j\le 100+x_{k}(100), \\
  & 0,    \mbox{otherwise}. \\
\end{aligned} \right.$$
The entries of the true coefficient matrices (${{H}_{01}}\in R_{+}^{20\times 200},{{H}_{02}}\in R_{+}^{20\times 150},{{H}_{03}}\in R_{+}^{20\times 300}$) were regarded as independent and identical Bernoulli variables with probability equal to $1/20$. Then we set the data matrices by ${{X}_{0I}}={{W}_{0}}{{H}_{0I}}+\mu E$, ($I = 1, 2, 3$), where $E$ was Gaussian noise and $\mu=2$. ${{\Theta }_{I}}$ and ${{R}_{IJ}}$ were generated as mentioned in section \ref{RThe}.

\begin{table*}[htbp]
  \centering
  \begin{threeparttable}[b]
  \caption{Performance comparison of update algorithms on synthetic dataset 3.}
  \label{tabsolverComparison3}
    \begin{tabular}{l|rrr|rrr|rrr|rrr}
    \hline
    \multirow{2}[2]{*}{} & \multicolumn{3}{c|}{Time (seconds)} & \multicolumn{3}{c|}{\#iteration} & \multicolumn{3}{c|}{Reconstruction error} & \multicolumn{3}{c}{AUC} \\ \hline
    Stop1 & $10^{-5}$ & $10^{-6}$ & $10^{-7}$ & $10^{-5}$ & $10^{-6}$ & $10^{-7}$ & $10^{-5}$ & $10^{-6}$ & $10^{-7}$ & $10^{-5}$ & $10^{-6}$ & $10^{-7}$ \\
    \hline
    MUR   & \textbf{6.01} & 19.67  & 71.31  & 82    & 268   & 934   & 1761985.02  & 1748184.66  & 1743369.37  & 0.60  & 0.63  & 0.64  \\
    PG    & 42.87  & 129.66  & 428.45  & 25    & 68    & 213   & \textbf{1744257.88} & 1741166.87  & \textbf{1740083.05} & \textbf{0.63} & \textbf{0.65} & \textbf{0.66} \\
    Ne    & 25.95  & 84.73  & 279.05  & 23    & 67    & 213   & 1744342.91  & \textbf{1741156.57} & 1740122.46  & \textbf{0.63} & \textbf{0.65} & \textbf{0.66} \\
    PANLS & 7.12  & \textbf{7.12} & \textbf{7.12} & 16    & 16    & 16    & 1750415.28  & 1750415.28  & 1750415.28  & 0.62  & 0.62  & 0.62  \\
    \hline
    Stop1+ &       &       & \multicolumn{1}{r}{} &       &       & \multicolumn{1}{r}{} &       &       & \multicolumn{1}{r}{} &       &       &  \\
    \hline
    MUR   & \textbf{5.50} & 18.31  & 67.64  & 79    & 260   & 936   & 1765436.16  & 1752533.34  & 1747422.15  & 0.61  & 0.63  & 0.65  \\
    PG    & 45.42  & 152.41  & 500.40  & 26    & 77    & 247   & 1748471.26  & \textbf{1744875.66} & \textbf{1743766.27} & \textbf{0.64} & 0.65  & \textbf{0.67} \\
    Ne    & 27.04  & 99.86  & 290.43  & 24    & 75    & 206   & \textbf{1748100.58} & 1744994.86  & 1744082.25  & \textbf{0.64} & \textbf{0.66} & \textbf{0.67} \\
    PANLS & 7.26  & \textbf{7.26} & \textbf{7.26} & 16    & 16    & 16    & 1750520.76  & 1750520.76  & 1750520.76  & 0.62  & 0.62  & 0.62  \\
    \hline
    \multicolumn{1}{r}{} & \multicolumn{3}{r|}{} & \multicolumn{3}{r|}{} & \multicolumn{3}{r|}{} & \multicolumn{3}{r}{} \\
    \hline
    Stop2 & $10^{-2}$ & $10^{-3}$ & $10^{-4}$ & $10^{-2}$ & $10^{-3}$ & $10^{-4}$ & $10^{-2}$ & $10^{-3}$ & $10^{-4}$ & $10^{-2}$ & $10^{-3}$ & $10^{-4}$ \\
    \hline
    PG    & 14.00  & 14.00  & 223.53  & 11    & 11    & 119   & 1753328.27  & 1753328.27  & 1740534.37  & 0.61  & 0.61  & 0.65  \\
    Ne    & 9.12  & 9.12  & 160.41  & 11    & 11    & 130   & \textbf{1751546.25} & \textbf{1751546.25} & \textbf{1740484.63} & \textbf{0.62} & \textbf{0.62} & \textbf{0.66} \\
    PANLS & \textbf{5.96} & \textbf{5.96} & \textbf{6.16} & 11    & 11    & 12    & 1755280.95  & 1755280.95  & 1754447.37  & 0.61  & 0.61  & 0.61  \\
    \hline
    Stop2+ &       &       &       &       &       &       &       &       &       &       &       &  \\
    \hline
    PG    & 14.57  & 217.65  & 217.65  & 11    & 108   & 108   & 1757552.21  & \textbf{1744415.18} & \textbf{1744415.18} & \textbf{0.62} & 0.66  & 0.66  \\
    Ne    & 9.17  & 158.46  & 158.46  & 11    & 119   & 119   & \textbf{1754502.24} & 1744452.77  & 1744452.77  & \textbf{0.62} & \textbf{0.67} & \textbf{0.67} \\
    PANLS & \textbf{6.60} & \textbf{6.60} & \textbf{6.99} & 11    & 11    & 12    & 1755219.16  & 1755219.16  & 1754096.35  & 0.61  & 0.61  & 0.61  \\
   \hline
    \end{tabular}
    \end{threeparttable}
  \label{tab:addlabel}%
\end{table*}%

\begin{table*}[htbp]
  \centering
  \begin{threeparttable}[b]
  \caption{Performance comparison of update algorithms on synthetic dataset 4.}
\label{tabsolverComparison4}
    \begin{tabular}{l|rrr|rrr|rrr|rrr}
    \hline
    \multirow{2}[2]{*}{} & \multicolumn{3}{c|}{Time (seconds)} & \multicolumn{3}{c|}{\#iteration} & \multicolumn{3}{c|}{Reconstruction error} & \multicolumn{3}{c}{AUC} \\ \hline
    Stop1 & $10^{-5}$ & $10^{-6}$ & $10^{-7}$ & $10^{-5}$ & $10^{-6}$ & $10^{-7}$ & $10^{-5}$ & $10^{-6}$ & $10^{-7}$ & $10^{-5}$ & $10^{-6}$ & $10^{-7}$ \\
    \hline
    MUR   & (4) 96.53 & (4) 158.80 & (4) 462.66 & 274   & 450   & 1300  & 7388732.03  & 7388323.61  & 7388133.55  & 0.97  & 0.97  & 0.97  \\
    PG    & 169.98  & 169.98  & 169.98  & 19    & 19    & 19    & 7389863.05  & 7389863.05  & 7389863.05  & 0.97  & 0.97  & 0.97  \\
    Ne    & 1293.50  & 3098.76  & 4597.90  & 72    & 173   & 258   & \textbf{7388388.82} & \textbf{7388071.50} & \textbf{7388041.37} & 0.97  & 0.97  & 0.97  \\
    PANLS & \textbf{41.21} & \textbf{41.21} & \textbf{41.21} & 19    & 19    & 19    & 7390354.04  & 7390354.04  & 7390354.04  & 0.97  & 0.97  & 0.97  \\
    \hline
    Stop1+ &       &       & \multicolumn{1}{r}{} &       &       & \multicolumn{1}{r}{} &       &       & \multicolumn{1}{r}{} &       &       &  \\
    \hline
    MUR   & (6) 89.80 & (6)130.78 & (6) 536.48 & 269   & 391   & 1585  & 7388720.29  & 7388417.05  & 7388306.94  & 0.97  & 0.97  & 0.97  \\
    PG    & 189.51  & 189.51  & 189.51  & 21    & 21    & 21    & 7389985.35  & 7389985.35  & 7389985.35  & 0.97  & 0.97  & 0.97  \\
    Ne    & 1407.41  & 2053.11  & 6027.49  & 82    & 120   & 353   & \textbf{7388495.89} & \textbf{7388322.89} & \textbf{7388154.20} & 0.97  & 0.97  & 0.97  \\
    PANLS & \textbf{39.54} & \textbf{39.54} & \textbf{39.54} & 19    & 19    & 19    & 7390545.25  & 7390545.25  & 7390545.25  & 0.97  & 0.97  & 0.97  \\
    \hline
           \multicolumn{1}{r}{} & \multicolumn{3}{r|}{} & \multicolumn{3}{r|}{} & \multicolumn{3}{r|}{} & \multicolumn{3}{r}{} \\
    \hline
     Stop2 & $10^{-2}$ & $10^{-3}$ & $10^{-4}$ & $10^{-2}$ & $10^{-3}$ & $10^{-4}$ & $10^{-2}$ & $10^{-3}$ & $10^{-4}$ & $10^{-2}$ & $10^{-3}$ & $10^{-4}$ \\
    \hline
    PG    & 951.22188 & 6083.72344 & 6083.72  & 96    & 617   & 617   & 7388252.66  & \textbf{7388026.86} & \textbf{7388026.86} & 0.97  & 0.97  & 0.97  \\
    Ne    & 1725.58  & 11310.29  & 11310.29  & 93    & 612   & 612   & \textbf{7388250.54} & \textbf{7388026.86} & \textbf{7388026.86} & 0.97  & 0.97  & 0.97  \\
    PANLS & \textbf{31.76} & \textbf{41.11} & \textbf{54.16} & 14    & 19    & 28    & 7400537.43  & 7390817.86  & 7390188.40  & 0.94  & 0.97  & 0.97  \\
    \hline
    Stop2+ &       &       & \multicolumn{1}{l}{} &       &       & \multicolumn{1}{l}{} &       &       & \multicolumn{1}{l}{} &       &       &  \\
   \hline
    PG    & 4048.95  & 4048.95  & 4048.95  & 437   & 437   & 437   & \textbf{7388215.39} & \textbf{7388215.39} & \textbf{7388215.39} & 0.97  & 0.97  & 0.97  \\
    Ne    & (5) 1718.61 & (5) 1718.61 & (5) 1718.61 & 101   & 101   & 101   & 7399093.15  & 7399093.15  & 7399093.15  & \textbf{0.98} & \textbf{0.98} & \textbf{0.98} \\
    PANLS & \textbf{29.63} & \textbf{39.70} & \textbf{61.04} & 14    & 19    & 33    & 7400044.34  & 7390545.25  & 7390303.49  & 0.94  & 0.97  & 0.97  \\
     \hline
    \end{tabular}
    \end{threeparttable}
  \label{tab:addlabel}%
\end{table*}%

\subsection{Synthetic dataset 4}
Finally, we simulated a big dataset by increasing the dimensions of $H_I$ to demonstrate the effectiveness of the algorithms for JMF. The true low-rank $=5$ and the ground truth basis matrix ${{W}_{0}}\in R_{+}^{500\times 5}$ was constructed with \emph{coph} =$0$ as follows,
\begin{equation}
  \mbox{$\displaystyle
  {{W}_{0}}[j,k]=\left\{ \begin{aligned}
  & 1,    1+x_{k}(100)\le j\le 100+x_{k}(100), \\
 & 0,    \mbox{otherwise}. \\
 \end{aligned} \right.
$}
\end{equation}
Meanwhile, three coefficient matrices (${{H}_{01}}\in R_{+}^{5\times 1200},{{H}_{02}}\in R_{+}^{5\times 1300},{{H}_{03}}\in R_{+}^{5\times 2000}$) were constructed with \emph{coph} =$0$,
\begin{equation}
  \mbox{$\displaystyle
{{H}_{01}}[j,k]=\left\{ \begin{aligned}
  & 1, 1+x_{j}(240) \le k\le 240+x_{j}(240), \\
 & 0,\mbox{otherwise}. \\
\end{aligned} \right.
$}
\end{equation}
\begin{equation}
  \mbox{$\displaystyle
{{H}_{02}}[j,k]=\left\{ \begin{aligned}
  & 1, 1+x_{j}(260) \le k\le 260+x_{j}(260), \\
 & 0, \mbox{otherwise}. \\
\end{aligned} \right.
$}
\end{equation}
\begin{equation}
  \mbox{$\displaystyle
{{H}_{03}}[j,k]=\left\{ \begin{aligned}
  & 1,    1+x_{j}(400) \le k\le 400+x_{j}(400), \\
 & 0,\mbox{otherwise}. \\
\end{aligned} \right.
$}
\end{equation}
We set the data matrices by ${{X}_{0I}}={{W}_{0}}{{H}_{0I}}+\mu E$ ($I = 1, 2, 3$), where $E$ was Gaussian noise and $\mu=3$. ${{\Theta }_{I}}$ and ${{R}_{IJ}}$ were generated as mentioned in section \ref{RThe}.

\subsection{Parameter selection}
\begin{figure}[htp]
\centerline{\includegraphics[width=0.48\textwidth]{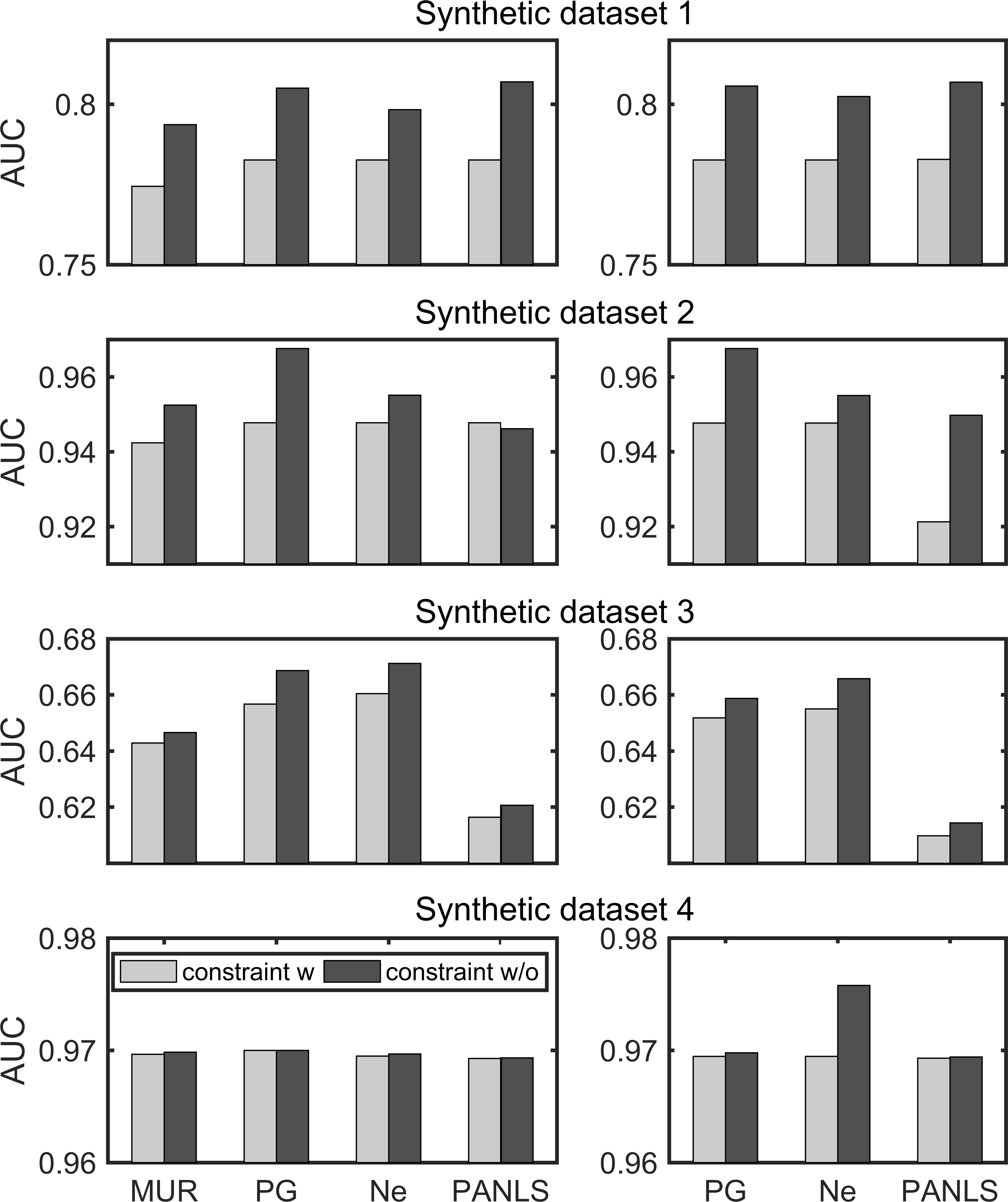}}
\caption{Comparison of the effectiveness in terms of AUC of MUR, PG, Ne and PANLS for JMF with or without constraints under various noise level on synthetic datasets. The AUC value is the average of $10$ realizations.} \label{Constraints}
\end{figure}

In the model, there are four parameters in total. Figure \ref{Constraints influence} shows that how the performance of JMF varies with respect to each parameter on the synthetic datasets. As we normalized each row of $H_I$ ($I = 1, 2,\dots ,N$) after each iteration, the performance is almost not influenced by the sparse constraint parameter value $\gamma_{2}$, while it is a bit susceptible to $\gamma_{1}$. The model is relatively robust to $\lambda_{1}$, $\lambda_{2}$ and $\gamma_{2}$. Generally, the constraints combined together improve the effectiveness (Figure \ref{Constraints}).

We selected parameters by grid searching and $\lambda_{1}$, $\lambda_{2}$, and $\gamma_{2}$ were selected from $[0.001, 0.01, 0.1, 1, 10, 100, 1000]$, while $\gamma_{1}$ was selected from $[10^{-6}, 10^{-5}, 10^{-4}, 0.001, 0.01]$. We run the algorithm with each group of parameters $3$ times and computed the average AUC value for each group of parameters. The best parameter combination was selected with relatively small reconstruction error and large AUC values. The performance of these models were evaluated by identifying the pattern of original data matrices for synthetic datasets, respectively. The performance of the results with constraints on all synthetic datasets are better than those without constraints, suggesting the importance of the regularized-network constraints. 

\subsection{Convergence and complexity analysis}
\begin{figure}[htp]
\centerline{\includegraphics[width=0.48\textwidth]{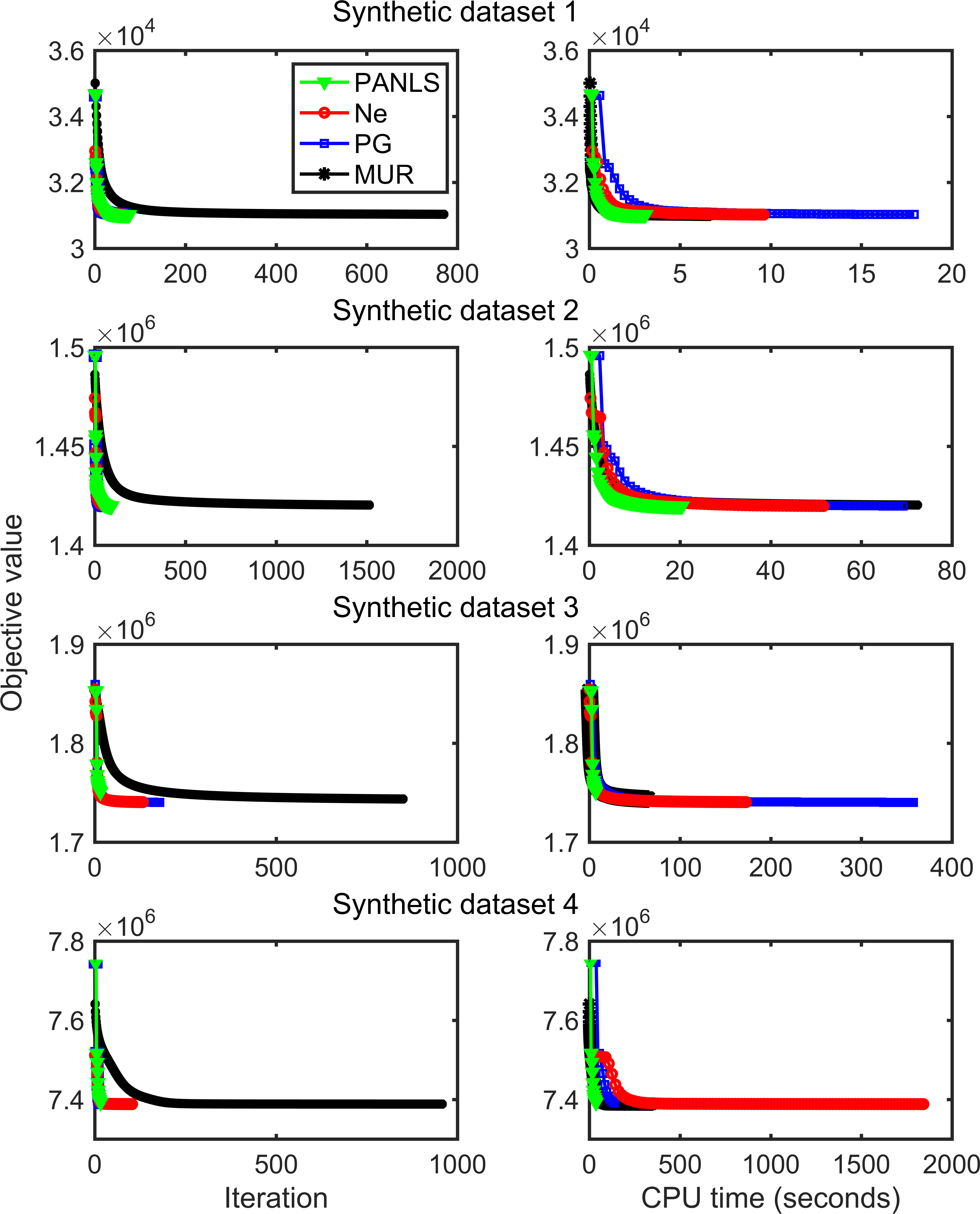}}
\caption{Average objective values with respect to iteration numbers and CPU time of PG, Ne and PANLS for JMF on synthetic datasets based on the objective-based criterion. The $y$-axis represents the objective value and $x$-axis represents the iteration number or CPU time (seconds).} \label{criterion 1}
\end{figure}

\begin{figure}[htp]
\centerline{\includegraphics[width=0.48\textwidth]{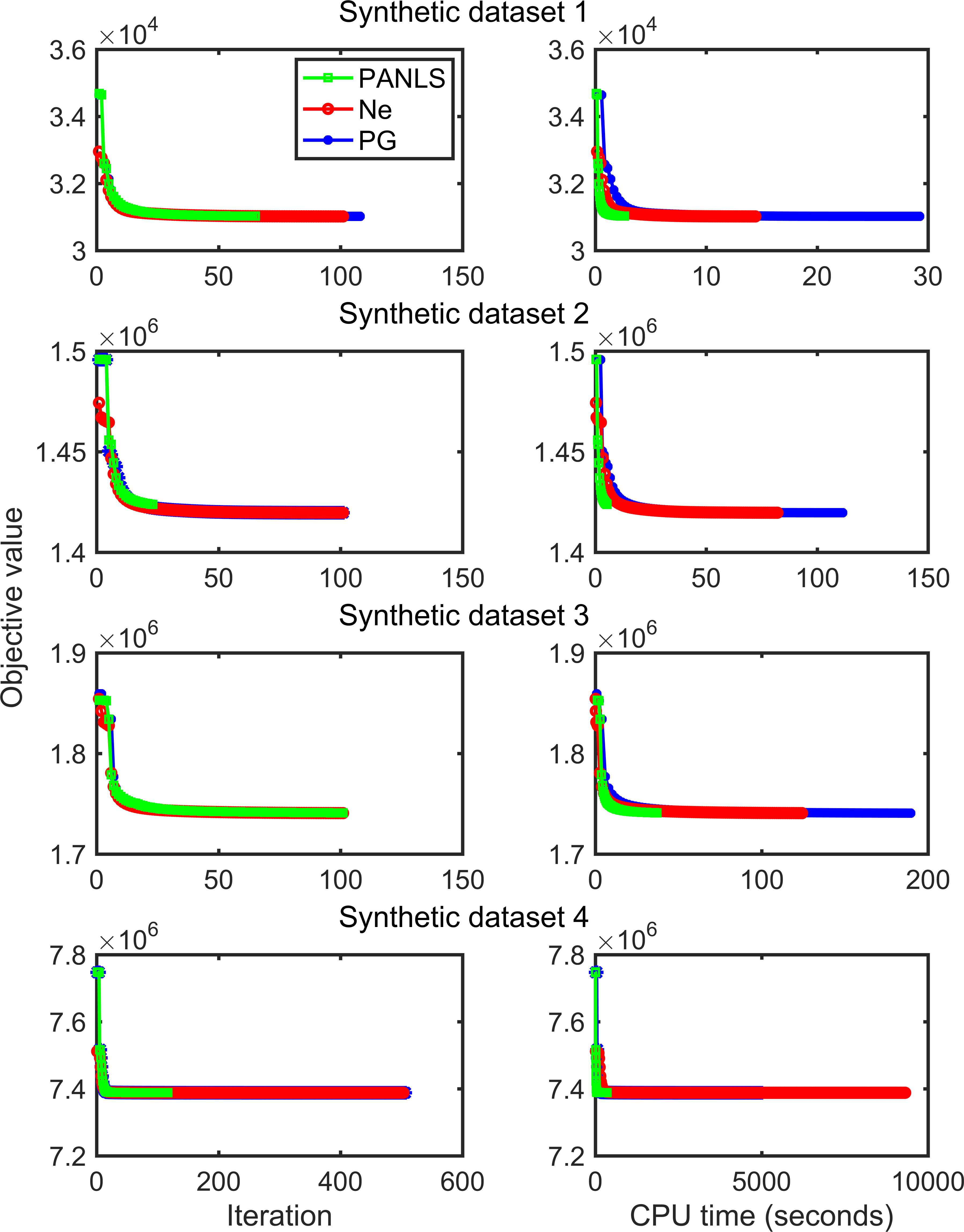}}
\caption{Average objective values with respect to iteration numbers and CPU time of PG, Ne and PANLS for JMF on synthetic datasets based on the gradient-based stop criterion. The $y$-axis represents the objective value and $x$-axis represents the iteration number or CPU time (seconds).} \label{criterion 2}
\end{figure}

Figure \ref{criterion 1} shows the average objective values versus iteration numbers or CPU times of the convergence curves of these four update rules for JMF on the synthetic datasets based on the first objective-based stop criterion defined in Eq. \ref{stop 1} and we set $\tau $ to $10^{-7}$. We can see that PG, Ne and PANLS decrease objective value sharply in each step and PANLS converges in less time to obtain an optimal solution than others. Figure \ref{criterion 2} shows the average objective values versus iteration numbers or CPU times of the convergence curves of PG, Ne and PANLS for JMF on the synthetic datasets based on the second gradient-based stop criterion defined in Eq. \ref{stop 2}. PANLS uses least time among them when select proper initial values and obtains a relatively good solution. From Tables 1-4, we can see that a good initial solution would make the MUR satisfies stop condition within $2000$ steps and other three algorithms stop within $200$ iterations.

We summarized the time complexity of one iteration round of the four update methods (Table \ref{tabtimeComplexity}). Such complexity of these methods are very comparative. However, PANLS converges in less time within each iteration round and in smaller iteration number (Figures 4, 5 and Tables 1-4), enabling it is the fastest one among these four algorithms.
\begin{table*}[htbp]
  \centering
   \begin{threeparttable}[b]
  \caption{Computational complexity for one round of each method}
  \label{tabtimeComplexity}%
    \begin{tabular}{rll}
    \hline
          & Update algorithms     & Time complexity \\
    \hline
    \rule{0pt}{10pt} & MUR & $O(mnr+{{n}^{2}}r+m{{r}^{2}}+n{{r}^{2}})$ \\
    \rule{0pt}{10pt} & PG & $O(mnr+{{n}^{2}}r+m{{r}^{2}}+n{{r}^{2}})+K\times O(tm{{r}^{2}}+t\sum_{I}{n_{I}^{2}}r+tn{{r}^{2}})$  \quad\quad\quad\quad\quad\quad\quad\quad\quad\quad\quad  \\
    \rule{0pt}{10pt} & NeNMF & $O(mnr+{{n}^{2}}r+m{{r}^{2}}+n{{r}^{2}})+K\times O(m{{r}^{2}}+n{{r}^{2}})$ \\
    \rule{0pt}{10pt} & PANLS & $O(mnr+{{n}^{2}}r+m{{r}^{2}}+n{{r}^{2}})+K\times O(m{{r}^{2}}+A)$\\
    \hline
    \end{tabular}%
    \begin{tablenotes}
    \item \small {$m$ is the row number of any data matrices, $r$ is the low rank, $n_{i}$ is the column number of the $i$th data matrix, and $n$ is the sum of $n_{i}$. $K$ is the inner iteration, and $t$ is the iteration of the line search procedure. As the time complexity of the inner computation of PANLS is hard to estimate, we represent it as A.}
    \end{tablenotes}
    \end{threeparttable}
  \label{tab:addlabel}%
\end{table*}%

As MUR cannot guarantee convergence, we just run MUR algorithm with the first stop criterion. We compared these various algorithms on the synthetic datasets (Table \ref{tabsolverComparison1}-\ref{tabsolverComparison4}). PG and Ne have relatively small reconstruction error and better AUC values. MUR only quickly decreases the objective value at the beginning, while PANLS has the fastest speed and almost competitive accuracy with Ne and PG.

\subsection{Application onto real data I}
RNA binding proteins (RBPs) control gene expression in post-transcription through splicing, transport, polyadenylation, RNA stability and so on. The interactions of protein-RNA are affected by various aspects. A recent study was designed to predict the protein-RNA interactions by integrating several datasets (RBP experimental data, gene function, RNA sequence and structure) \cite{stravzar2016orthogonal}. We applied JMF/L model to predict protein-DNA interactions. As there was no information of must-links and not-links, the parameters $\lambda_{1}$ and $\lambda_{2}$ were set to zeros. The left two parameters ($\gamma_{1}$ and $\gamma_{2}$) were searched by grid search in the range $[10^{-6}, 10^{-5}, \dots , 10^{-3}]$ and $[10^{-3}, 10^{-2}, \dots , 10^{3}]$. The factorization rank was set to $10$ as suggested in \cite{stravzar2016orthogonal} and run JMF with the same three randomly initial values for algorithms. We used five fold cross-validation on the training set of $5000$ positions to choose hyper-parameters on each RBP experiment. Then we evaluated the performance of the algorithms in terms of AUC and running time. We set the stop tolerance equal to $10^{-8}$. Generally, the result illustrates that Ne has the best AUC, while PANLS uses least time (Table \ref{result_I}).

\begin{table}[htbp]
  \centering
  \footnotesize  
  \setlength{\tabcolsep}{4.5pt}
  \begin{threeparttable}[b]
  \caption{Prediction performance in term of AUC and running time using JMF with four update algorithms. }
  \label{result_I}%
    \begin{tabular}{lrrrr|rrrr}
    \hline
                 & \multicolumn{4}{c|}{AUC}      & \multicolumn{4}{c}{Time (seconds)} \\
    \hline
    \multicolumn{1}{r}{Protein} & MUR   & PG    & Ne    & PA & MUR   & PG    & Ne    & PA \\
    \hline
    $1$ Ago/EIF. & 0.88  & 0.88  & 0.88  & 0.88                  & 721   & 624    & 640  & \textbf{326} \\
    $2$ Ago2M.   & 0.70  & \textbf{0.72} & 0.69  & 0.69          & 1400  & 741    & 565  & \textbf{365} \\
    $3$ Ago2     & \textbf{0.82} & 0.81  & \textbf{0.82} & 0.81  & 1317  & 694    & 695  & \textbf{341} \\
    $4$ Ago2     & 0.81  & \textbf{0.83} & \textbf{0.83} & 0.82  & 1603  & 621    & 601  & \textbf{341} \\
    $5$ Ago2     & 0.71  & 0.71  & \textbf{0.72} & 0.69          & 1235  & 665    & 597  & \textbf{416} \\
    $6$ eIF4AIII & 0.92  & \textbf{0.93} & \textbf{0.93} & \textbf{0.93} & 1744   & 815  & 609  & \textbf{560} \\
    $7$ eIF4AIII & 0.90  & 0.90  & 0.90  & 0.90                  & 1235  & 866    & 904  & \textbf{368} \\
    $8$ ELAVL1   & 0.84  & 0.83  & 0.82  & \textbf{0.85}         & 641   & 770    & 388  & \textbf{359} \\
    $9$ ELAVL1M. & 0.72  & \textbf{0.73} & 0.72  & 0.72          & 1198  & 926    & 449  & \textbf{415} \\
    $10$ ELAVL1A  & 0.92  & \textbf{0.93} & \textbf{0.93} & \textbf{0.93}& 805    & 608  & 403  & \textbf{399} \\
    $11$ ELAVL1   & 0.95  & 0.95  & 0.95  & 0.95                 & 790   & 659    & \textbf{404} & 441  \\
    $12$ ESWR1    & \textbf{0.83} & 0.82  & \textbf{0.83} & 0.82 & 899   & 714    & 495  & \textbf{316} \\
    $13$ FUS      & 0.68  & \textbf{0.74} & 0.71  & 0.65         & 848   & 557    & 507  & \textbf{306} \\
    $14$ Mut FUS  & 0.93  & \textbf{0.94} & \textbf{0.94} & 0.94 & 1173  & 696    & 565  & \textbf{376} \\
    $15$ IGF2.1-3 & 0.92  & \textbf{0.93} & \textbf{0.93} & \textbf{0.93} & 834   & 720  & 626  & \textbf{421} \\
    $16$ hnRNPC   & 0.73  & 0.72  & \textbf{0.86} & 0.73  & \textbf{138} & 263    & 244  & 152  \\
    $17$ hnRNPC   & 0.76  & 0.72  & \textbf{0.93} & 0.89  & \textbf{95}  & 314    & 212  & 142  \\
    $18$ hnRNPL   & 0.73  & \textbf{0.74} & \textbf{0.74} & \textbf{0.74} & 300   & 548  & \textbf{202} & 333  \\
    $19$ hnRNPL   & 0.61  & \textbf{0.62} & \textbf{0.62} & 0.59  & 990   & 635   & 563  & \textbf{330} \\
    $20$ hnRNPLl. & 0.65  & 0.65  & \textbf{0.66} & \textbf{0.66} & 1223  & 785   & 534  & \textbf{376} \\
    $21$ MOV10    & 0.95  & 0.95  & \textbf{0.96} & 0.95          & 801   & 515   & 467  & \textbf{392} \\
    $22$ Nsun2    & \textbf{0.78} & 0.77  & 0.77  & 0.77          & 1098  & 540   & 391  & \textbf{380} \\
    $23$ PUM2     & 0.90  & 0.91  & 0.91  & \textbf{0.92}         & 444   & 646   & \textbf{304} & 327  \\
    $24$ QKI      & 0.60  & 0.63  & \textbf{0.73} & 0.61          & 183   & 383   & 237  & \textbf{122} \\
    $25$ SRSF1    & 0.85  & 0.85  & 0.85  & 0.85                  & 1028  & 704   & 613  & \textbf{292} \\
    $26$ TAF15    & 0.88  & \textbf{0.89} & \textbf{0.89} & \textbf{0.89} & 1235  & 651  & 753   & \textbf{389} \\
    $27$ TDP-43   & 0.73  & 0.69  & 0.72  & \textbf{0.77}         & 430   & 322   & \textbf{221} & 270  \\
    $28$ TIA1     & \textbf{0.91} & \textbf{0.91} & 0.89  & \textbf{0.91} & 1130  & 841  & 534   & \textbf{474} \\
    $29$ TIAL1    & 0.81  & 0.81  & 0.82  & \textbf{0.83}         & 940   & 929   & 706  & \textbf{536} \\
    $30$ U2AF2    & \textbf{0.74} & 0.70  & \textbf{0.74} & 0.72  & 966   & 1044  & 592  & \textbf{347} \\
    $31$ U2AF2    & 0.67  & \textbf{0.71} & 0.68  & 0.70          & 1254  & 729   & 667  & \textbf{378} \\
    \hline
    \end{tabular}%
     \begin{tablenotes}
    \item \small {PA indicates PANLS.}
    \end{tablenotes}
    \end{threeparttable}
\end{table}%

\begin{table*}
\begin{threeparttable}[b]
  \caption{Summary of $15$ modules detected by JMF on breast cancer data}
  \label{table apllication II}
    \begin{tabular}{lllllllll}
    \hline
    No.   & Ge    & Mi(Ge) & Me(Ge) & Ln(Ge) & Oa    & Ob    & Oc    & Selected over-represented functional sets \\
    \hline
    26    & 487   & 0     & 489   & 13    & 0     & 13    & 22*   & Collagen formation; NCAM signaling \\
    40    & 674   & 0     & 461   & 2873  & 0     & 14    & 190*  & Cell Cycle; Mitotic M-M/G1 phases \\
    45    & 426   & 0     & 439   & 2895  & 0     & 18**  & 71    & O-Glycan biosynthesis \\
    48    & 649   & 26    & 482   & 2877  & 3     & 23**  & 107   & NGF signaling; Adipocytokine signaling; PPAR signaling \\
    55    & 688   & 56    & 511   & 4     & 4     & 21    & 2**   & Cytokine Signaling in Immune system; Adaptive Immune System \\
    66    & 434   & 36    & 376   & 16    & 4     & 8     & 3*    & Response to the detection of DNA damage \\
    71    & 455   & 0     & 440   & 37    & 0     & 22**  & 2     & Genes up-regulated in the luminal B subtype of breast cancer \\
    92    & 443   & 0     & 358   & 13    & 0     & 17**  & 3**   & Genes up-regulated in basal subtype of breast cancer samles. \\
    95    & 647   & 13    & 505   & 2877  & 1     & 14**  & 127   & Metabolism of lipids and lipoproteins; Cholesterol biosynthesis \\
    112   & 663   & 36    & 502   & 25    & 1     & 20    & 4*    & ECM-receptor interaction; PDGF signaling \\
    137   & 634   & 0     & 486   & 2874  & 0     & 13    & 195*  & Cell Cycle; Mitotic; DNA Replication \\
    140   & 558   & 0     & 415   & 16    & 0     & 18**  & 2*    & Genes up-regulated in a breast cancer cell line resistant to tamoxifen \quad\quad\quad\quad \\
    143   & 481   & 0     & 397   & 37    & 0     & 17    & 7***  & Cell development; Cell differentiation \\
    147   & 598   & 0     & 462   & 28    & 0     & 19*   & 3     & Genes down-regulated in the luminal B subtype of breast cancer \\
    164   & 415   & 17    & 334   & 33    & 2     & 16*   & 5**   & ECM-receptor interaction; Focal adhesion \\
    \hline
    \end{tabular}%
    \begin{tablenotes}
    \item  \small {No.: the index of the md-module. Ge: number of genes in GE dimension. Me(Ge): number of DM markers adjacent genes. Mi(Ge): number
of miRNAs targeting genes. Ln(Ge): number of lncRNAs targeting genes. Oa: overlap between gene set and DM markers adjacent gene set; Ob: overlap between gene set and miR
target gene set. Oc: overlap between gene set and lncRNAs target gene set. Where $* (0.05,0.1)$, $** (0.01,0.05)$ and $*** (0,0.01)$ indicate the $p$-value for the hypergeometric test, respectively.}
   \end{tablenotes}
   \end{threeparttable}
  \label{tab:addlabel}%
\end{table*}%

\subsection{Application onto real data II}
With more understanding of biological mechanism and the development of technology, various kinds of biological data has been generated. Cancer is a complex disease which influenced by both environmental and genetical factors including gene expression, DNA methylation, microRNA (miRNA) and lncRNAs ad so on. We applied JMF framework with PANLS to explore the pathogenic mechanism in breast cancer. We downloaded RNA-seq gene expression data (GE), miRNA expression data (ME), and DNA methylation data (DM) of breast cancer from TCGA on 2016-01-28. And we downloaded lncRNA expression data (LE) from TRANRIC database \cite{li2015tanric}. There were $379$ samples shared these datasets in total. We scaled each data matrix by dividing the median element. We searched $r$ from $50$ to $250$ increased by $50$ and $r = 200$ was selected with the least collinearity of any pair columns of $W$ and the largest mean correlation coefficients value between the original data and reconstructed data. We assigned feature $i$ (gene, miRNA, DNA methylation, and lncRNA) to module $j$ if the $z$-score of $h_{i}^{I}$ exceeds a threshold $T$ (e.g., $T$ = $1.5$).

\begin{figure}[htp]
\centerline{\includegraphics[width=.29\textwidth]{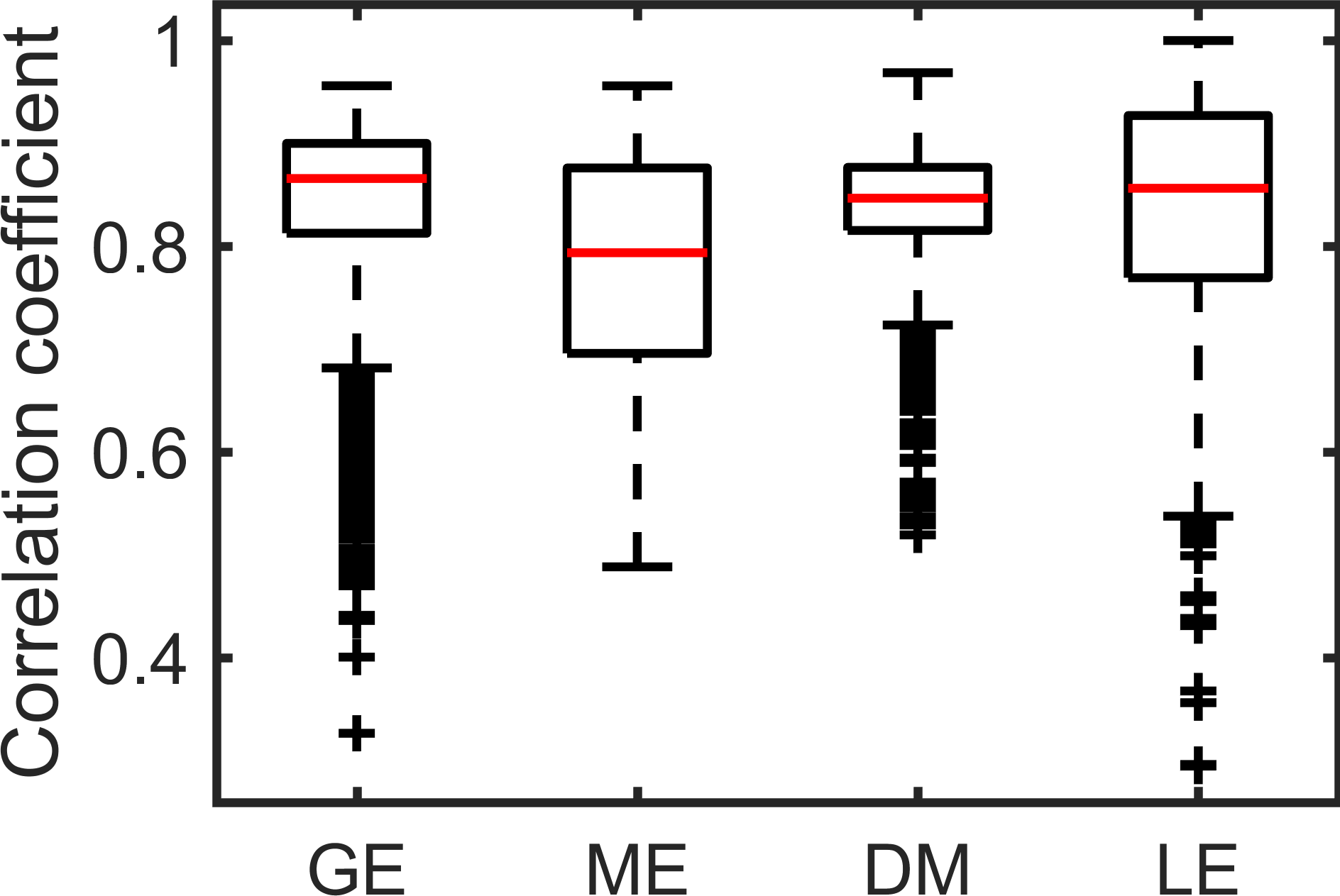}}
\caption{Sample-wise correlations of original and reconstructed multi-view datasets} \label{Real Dataset $2$A}
\end{figure}

The average correlations of the original and reconstructed gene, miRNA, methylation, and lncRNA profiles were $0.88$, $0.89$, $0.87$ and $0.79$ respectively, indicating that the dimension reduction captures the most information of original data (Figure \ref{Real Dataset $2$A}). To test the vertical associations of these $200$ modules, we randomly permuted modules with the same dimensions $1000$ times. $161$ modules have significant higher Pearson's correlation coefficients between any two of gene expression, miRNA expression, DNA methylation and lncRNA expression dimensions with $P$-value $<$ 0.05.

We downloaded GO biological process terms from MSigDB (Molecular Signatures Database) \cite{subramanian2005gene}, and detected enriched biological processes by Fisher's exact test on member genes in gene expression view, genes directly adjacent to member DNA methylation in DNA methylation view, and target genes by member miRNA and lncRNA in these two views, respectively. $192$ modules have at least one common enriched biological process with $P$-value $<0.05$. Therefore, the genes associated with these four views are functionally homogenous. Table \ref{table apllication II} shows 15 modules detected by JMF, which have overlapping genes between different dimensions within the same modules. However, there is little overlap between miRNA targeted genes and genes from other views. This might because miRNA usually repress gene expression, while module information identified by NMF is often superimposed with non-negative value. In module 126, \emph{CTHRC1} is the overlapping gene from all views, and \emph{CTHRC1} up-regulation is tightly associated with breast cancer carcinogenesis \cite{kharaishvili2011collagen,kim2013collagen}. hsa-let-7b-5p is specific to module 126 in miRNA expression view. Moreover, hsa-let-7b has been reported to be associated with metastasis of breast cancer \cite{marino2014microrna}.
\emph{CTHRC1} expression has been found dramatically and aberrantly up-regulated in the vast majority of human cancers including breast cancer \cite{tang2006aberrant}. By integrating multi-views datasets, many functional pathways can be detected which are not detected from a single-view data. For example, module 50 tends to be enriched in ``glycosaminoglycan biosynthesis keratan sulfate". Previous studies have shown that the glycosaminoglycan is strongly associated with the cell invasion and cell motility of breast cancer \cite{hamilton2007hyaluronan}.
Therefore, JMF is an effective tool to discover synergistic mechanism in breast cancer.

\begin{figure}[htp]
\centerline{\includegraphics[width=0.49\textwidth]{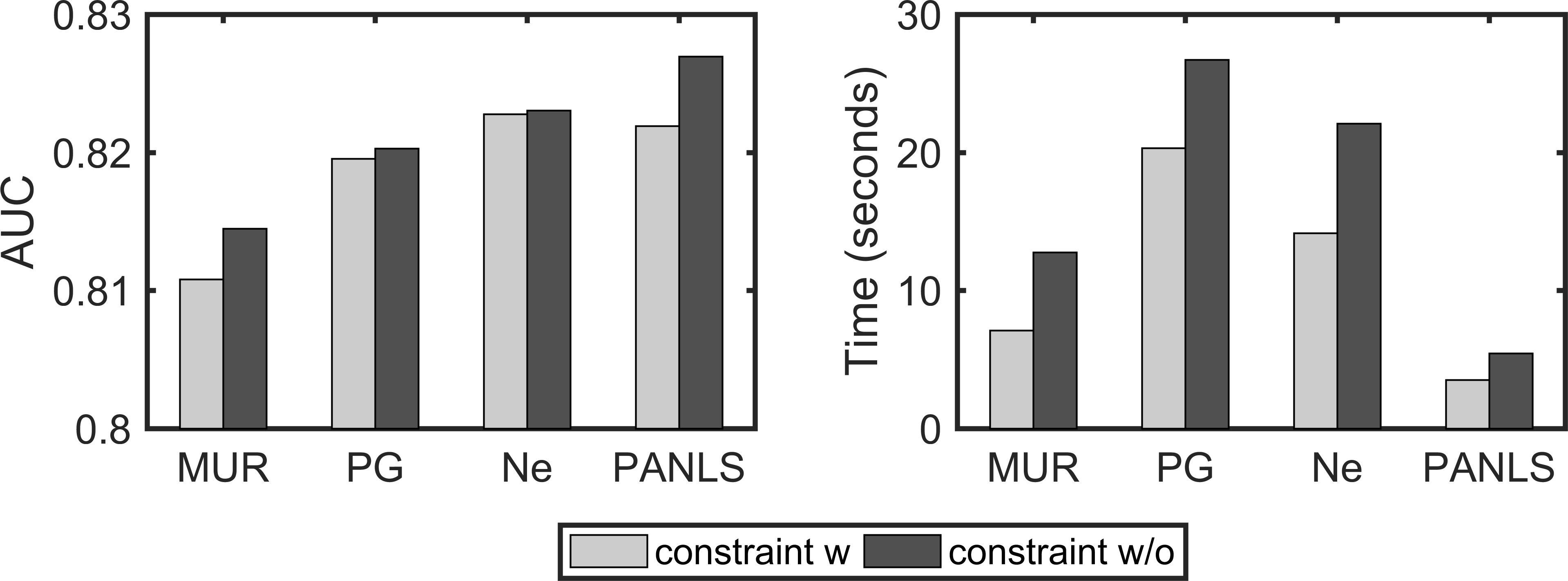}}
\caption{The classification performance of JMF on document clustering problem in terms of AUC and running time, respectively.} \label{Real Dataset $3$}
\end{figure}

\subsection{Application onto real data III}
In this section, we applied JMF onto a real-world multi-view document clustering task. We downloaded the dataset used in a previous study \cite{greene2009matrix}. There are $416$ distinct news stories from BBC, Reuters and The Guardian. These news were manually annotated with one or more of the six topical labels: business, entertainment, health, politics, sport, technology. We focused on the disjoint categories and used non-overlapping annotated topic classes, which were based on dominant topic for each story. Therefore, there were three matrices with $4750$ rows (words) and $313$, $254$, $266$ columns (news) from three news sources.

We constructed constraint matrices by the Pearson's correlation coefficients matrices on each view and between view data matrices. We ran each algorithm $30$ times and evaluated the performance with the average value of AUC and the average time of the algorithms with the stop tolerance equaling to $10^{-7}$. The results demonstrate that JMF with network-regularized constraints generally have better or competitive performance than those of without. Generally, PANLS and Ne and PG show distinct better performance than MUR. Moreover, among the four update algorithms, PANLS used least time, demonstrating its superior efficiency (Figure \ref{Real Dataset $3$}).

\section{Conclusion}\label{Conclusion}
We have presented a general framework for integrating prior network information with multi-view data sources simultaneously. The framework is flexible to identify multi-view linked patterns or make predictions. The performance of four widely used update methods were compared for solving each subproblem of the framework. Numerical results demonstrate that a new active method PANLS proposed recently is an attractive approach with high computational speed and competitive performance on synthetic and real datasets. Moreover, more informative linked-patterns are detected by JMF through integrating multi-view data, and prior knowledge represented by network-regularized constraints distinctly improve the prediction performance.

\appendices
\section{Update algorithms for solving JMF}
\begin{myalg}[\textbf{Solving JMF by MUR}] \label{alg_1}
\begin{algorithmic}[1]
\State{Initialize $W$, $H_I$ ($I=1,2,\ldots ,N$) with non-negative values, scale the columns of $H_I$
   to unit norm and set the iteration index $t = 0$.}
\State{Fix $H_I$ ($I=1,2,\ldots ,N$), update $W$ in problem Eq. \ref{update W} according to Eq. \ref{update}.}
\State{Fix $W$, update $H_I$ ($I=1,2,\ldots ,N$) in problem Eq. \ref{update H_I} according to Eq. \ref{update}.}
\State{Let $t\leftarrow t+1$, repeat \textbf{Step 2--3}
until convergence criteria are satisfied.}
\end{algorithmic}
\end{myalg}

\begin{myalg}[\textbf{Solving JMF by PG}] \label{alg_2}
\begin{algorithmic}[1]
\State{Initialize $W$, $H_I$ ($I=1,2,\ldots ,N$) with non-negative values, scale the columns of $H_I$ to unit norm and set the outer iteration index $t = 0$, $\sigma = 0.01$, ${{\alpha }_{0}} = 1$, $\beta = 0.1$. Given predefined tolerance $tol = 10^{-6}$ and inner iteration $K = 500$.}
\State{Fix $H_I$ ($I=1,2,\ldots ,N$), update $W$ on the constrained problem Eq. \ref{update W} by the following steps,

\begin{enumerate}[(2a)]
 \item Compute $proj = {{\left\| \nabla F({{W}^{k}})\left[ \nabla F({{W}^{k}})||{{W}^{k}}>0 \right] \right\|}_{2}}$, if $proj < tol$, go to \textbf{Step 3}, else go to \textbf{Step 2b}. Set the inner iteration index $k=0$.
 \item Compute ${{W}^{k+1}}=P({{W}^{k}}-{{\alpha }_{k}}\nabla F({{W}^{k}}))$,
  where ${\alpha }_{k}={\beta }^{t_{k}}$, and $t_{k}$ is the first non-negative integer for which satisfies
  {\small$$\begin{aligned}
  &(1-\sigma )\left\langle \nabla F({{W}^{k}}) \right.,\left. {{W}^{k+1}}-{{W}^{k}} \right\rangle \\
  &+\frac{1}{2}\left\langle {{W}^{k+1}}-{{W}^{k}} \right.,\left. {{Q}_{W}}({{W}^{k+1}}-{{W}^{k}}) \right\rangle \le 0.
  \end{aligned}$$}
  \item Let $k\leftarrow k+1$, repeat \textbf{Step 2a--2c} until $k>K$.
 \end{enumerate}}

\State{Fix $W$, update $H_I$ ($I=1,2,\ldots ,N$) on the constrained problem Eq. \ref{update H} by the following steps,
\begin{enumerate}[(3a)]
\item Compute $proj = {{\left\| \nabla F({{H_{I}}^{k}})\left[ \nabla F({{H_{I}}^{k}})||{{H_{I}}^{k}}>0 \right] \right\|}_{2}}$, if $proj < tol$, go to \textbf{Step 4}, else go to \textbf{Step 3b}. Set the inner iteration index $k=0$.
\item Compute $H_{I}^{k+1}=P(H_{I}^{k}-{{\alpha }_{k}}\nabla F(H_{I}^{k}))$,
  where ${\alpha }_{k}={\beta }^{t_{k}}$, and $t_{k}$ is the first non-negative integer for which satisfies
  {\small$$\begin{aligned}
  &(1-\sigma )\left\langle \nabla F(H_{I}^{k}) \right.,\left. H_{I}^{k+1}-H_{I}^{k} \right\rangle \\
  &+\frac{1}{2}\left\langle H_{I}^{k+1}-{{H}^{k}} \right.,\left. {{Q}_{{{H}_{I}}}}(H_{I}^{k+1}-H_{I}^{k}) \right\rangle \le 0.
   \end{aligned}$$}
\item Let $k\leftarrow k+1$, repeat \textbf{Step 3a--3c} until $k>K$.
 \end{enumerate}}
\State{Let $t\leftarrow t+1$, repeat \textbf{Step 2--3}
until convergence criteria are satisfied.}
\end{algorithmic}
\end{myalg}

\begin{myalg}[\textbf{Solving JMF by Ne}] \label{alg_3}
\begin{algorithmic}[1]
\State{Initialize $W$, $H_I$ ($I=1,2,\ldots ,N$) with non-negative values, scale the columns of $H_I$ to unit norm and set the outer iteration index $t = 0$, ${{\alpha }_{0}} = 1$, predefined tolerance $tol = 10^{-6}$, and inner iteration $K = 500$.}
\State{Fix $H_I$ ($I=1,2,\ldots ,N$), update $W$ on the constrained problem Eq. \ref{update W} by the following steps,

\begin{enumerate}[(2a)]
\item Compute $proj = \nabla_{W}^{P}F(W^{k})$, if $proj < tol$, go to \textbf{Step 3}, else go to \textbf{Step 2b}. Set the inner iteration index $k=0$.
\item Compute ${{L}_{W}}=2{{\left\| \sum\limits_{I}{H_{I}^{k}{{(H_{I}^{k})}^{T}}+{{\gamma }_{1}}I} \right\|}_{2}}$, update
  $W_{k}$, $\alpha_{k+1}$ and $Y_{k+1}$ with
  $${{W}^{k}}=P({{Y}^{k}}-\frac{1}{{{L}_{W}}}{{\nabla }_{W}}F({{W}^{k}},{{Y}^{k}})),$$
  $${{\alpha }_{k+1}}=\frac{1+\sqrt{4\alpha _{k}^{2}+1}}{2},$$
  $${{Y}^{k+1}}={{W}^{k}}+\frac{{{\alpha }_{k}}-1}{{{\alpha }_{k+1}}}({{W}^{k}}-{{W}^{k-1}}).$$
\item Let $k\leftarrow k+1$, repeat \textbf{Step 2a--2c} until $k>K.$
\end{enumerate}}

\State{Fix $W$, update $H_I$ ($I=1,2,\ldots ,N$) on the constrained problem Eq. \ref{update H} by the following steps,

\begin{enumerate}[(3a)]
\item Compute $proj = \nabla_{H_{I}}^{P}F(H_{I}^{k})$, if $proj < tol$, go to \textbf{Step 4}, else go to \textbf{Step 3b}. Set the inner iteration index $k=0$.
\item Compute ${{L}_{{{H}_{I}}}}=2{{\left\| {{W}^{T}}W+{{\gamma }_{2}}\text{1}{{\text{1}}^{T}} \right\|}_{2}}+{{\lambda }_{1}}{{\left\| \sum\limits_{s}{\Theta _{I}^{s}+{{(\Theta _{I}^{s})}^{T}}} \right\|}_{2}}$,  update ${H}_{I}^{k}$, $\beta_{k+1}$ and ${YH_{I}}^{k+1}$ with
  $$H_{I}^{k}=P({{YH_{I}}^{k}}-\frac{1}{{{L}_{{{H}_{I}}}}}{{\nabla }_{{{H}_{I}}}}F(H_{I}^{k},{{YH_{I}}^{k}})),$$
  $${{\beta }_{k+1}}=\frac{1+\sqrt{4\beta _{k}^{2}+1}}{2},$$
  $${{YH_{I}}^{k+1}}=H_{I}^{k}+\frac{{{\beta }_{k}}-1}{{{\beta }_{k+1}}}(H_{I}^{k}-H_{I}^{k-1}).$$
\item Let $k\leftarrow k+1$, repeat \textbf{Step 3a--3c} until $k>K.$
\end{enumerate}}
\State{Let $t\leftarrow t+1$, repeat \textbf{Step 2--3}
until convergence criteria are satisfied.}
\end{algorithmic}
\end{myalg}

\begin{myalg}[\textbf{Solving JMF by PANLS}] \label{alg_4}
\begin{algorithmic}[1]
\State{Initialize $W$, $H_I$ ($I=1,2,\ldots ,N$) with non-negative values, scale the columns of $H_I$
   to unit norm and set the iteration index $t = 0$, and inner iteration $K = 500$. Set parameters $\eta = 0.1$, $\alpha = 1$, $\beta = 0.1$, $\rho = 0.5$, $n_{1} = 2$, $n_{2} = 1$, $\tau = 10^{-7}$, and predefined tolerance $tol = 10^{-6}$.}
\State{Fix $H_I$ ($I=1,2,\ldots ,N$), update $W$ on the constrained problem Eq. \ref{update W PANLS} by the following steps,

  Compute ${{\left\| {{\nabla }^{P}}F(W) \right\|}_{F}}$, while it exceeds $tol$ and the inner iteration index $k < K$, repeat:
  \begin{enumerate}[(2a)]
  \item Execute PG step to obtain $W^{k+1}$ from $W^{k}$, if ${{\left\| {{g}_{\mathcal{I}}}(W) \right\|}_{F}}<\eta {{\left\| {{\nabla }^{P}}F(W) \right\|}_{F}}$,
  where
  $${{({{g}_{\mathcal{I}}}(W))}_{ij}}=\left\{ \begin{matrix}
   {{({{\nabla }_{W}}F(W))}_{ij}},\text{  if }{{W}_{ij}}>0,  \\
            0,          \text{ if }{{W}_{ij}}=0\text{ }.  \\
\end{matrix} \right.$$
set  $\eta =\rho \eta $.
\item Else if the number of iterations in the loop exceeds $n_{1}$, then go to Step 2c.
\item Execute the unconstrained conjugate gradient method to obtain $W^{k+1}$ from $W^{k}$, if ${{\left\| {{g}_{\mathcal{I}}}(W) \right\|}_{F}}<\eta {{\left\| {{\nabla }^{P}}F(W) \right\|}_{F}}$, go to \textbf{Step 2a}.
    \item Else if $\mathcal{U}({W^{k}})\ne \varnothing $ and $0<\left| \mathcal{A}(W^{k+1}) \right|-\left| \mathcal{A}(W^{k}) \right|\le {{n}_{2}}$, then go to \textbf{Step 2a},
    where $U({{W}^{k}})=\{i:\left| {{g}_{i}}({{W}^{k}}) \right|\ge {{\left\| {{\nabla }^{P}}F({{W}^{k}}) \right\|}^{\alpha }}\text{and }W_{i}^{k}\ge {{\left\| {{\nabla }^{P}}F({{W}^{k}}) \right\|}^{\beta }}\}$.
    \item Else if $\left| \mathcal{A}(W^{k+1}) \right|>\left| \mathcal{A}(W^{k}) \right|+{{n}_{2}}$, restart the conjugate gradient method with the reduced dimension $\left| \mathcal{I}(W^{k+1}) \right|$ at $W^{k+1}$.
        \end{enumerate} }

\State{Fix $W$, update $H_I$ ($I=1,2,\ldots ,N$) on the constrained problem Eq. \ref{update H_I PANLS} by the following steps,

  Compute ${{\left\| {{\nabla }^{P}}F(W) \right\|}_{F}}$, while it exceeds $tol$ and the inner iteration index $k < K$, repeat:
  \begin{enumerate}[(3a)]
  \item Execute PG step to obtain ${H_{I}}^{k+1}$ from ${H_{I}}^{k}$, if
  $${{\left\| {{g}_{\mathcal{I}}}(H_{I}) \right\|}_{F}}<\eta {{\left\| {{\nabla }^{P}}F(H_{I}) \right\|}_{F}},$$
  where
  $${{({{g}_{\mathcal{I}}}(H_{I}))}_{ij}}=\left\{ \begin{matrix}
   {{({{\nabla }_{H_{I}}}F(H_{I}))}_{ij}}, \text{  if }{{H_{I}}_{ij}}>0,  \\
   \text{         0           , if }{{H_{I}}_{ij}}=0.  \\
\end{matrix} \right.$$
 set $\eta =\rho \eta $.
\item Else if the number of iterations in the loop exceeds $n_{1}$, then go to \textbf{Step 2c}.
\item Execute the unconstrained conjugate gradient method to obtain ${H_{I}}^{k+1}$ from ${H_{I}}^{k}$, if ${{\left\| {{g}_{\mathcal{I}}}(H_{I}) \right\|}_{F}}<\eta {{\left\| {{\nabla }^{P}}F(H_{I}) \right\|}_{F}}$, go to \textbf{Step 2a}.
    \item Else if $\mathcal{U}({H_{I}}^{k})\ne \varnothing $ and $0<\left| \mathcal{A}({H_{I}}^{k+1}) \right|-\left| \mathcal{A}({H_{I}}^{k}) \right|\le {{n}_{2}}$, then go to \textbf{Step 2a}, where $U(H_{I}^{k})=\{i:\left| {{g}_{i}}(H_{I}^{k}) \right|\ge {{\left\| {{\nabla }^{P}}F(H_{I}^{k}) \right\|}^{\alpha }}\text{and }({{H}_{I}})_{i}^{k}\ge {{\left\| {{\nabla }^{P}}F(H_{I}^{k}) \right\|}^{\beta }}\}$.
    \item Else if $\left| \mathcal{A}({H_{I}}^{k+1}) \right|>\left| \mathcal{A}({H_{I}}^{k}) \right|+{{n}_{2}}$, restart the conjugate gradient method with the reduced dimension $\left| \mathcal{I}({H_{I}}^{k+1}) \right|$ at ${H_{I}}^{k+1}$.
        \end{enumerate}}
\State{Let $t\leftarrow t+1$, repeat \textbf{Step 2--3}
until convergence criteria are satisfied.}
\end{algorithmic}
\end{myalg}



%

\end{document}